\documentclass{article}
\usepackage{arxiv}

\usepackage[utf8]{inputenc} % allow utf-8 input
\usepackage[T1]{fontenc}    % use 8-bit T1 fonts
\usepackage{hyperref}       % hyperlinks
\usepackage{url}            % simple URL typesetting
\usepackage{booktabs}       % professional-quality tables
\usepackage{amsfonts}       % blackboard math symbols
\usepackage{nicefrac}       % compact symbols for 1/2, etc.
\usepackage{microtype}      % microtypography
\usepackage{lipsum}         % for filler text
\usepackage{graphicx}       % figures
\usepackage{amsmath,amssymb}
\usepackage{textcomp}
\usepackage{wrapfig}
\usepackage{multirow}
\usepackage{subcaption}     % for subfigures (recommended)
\usepackage{algorithm}      % for algorithms
\usepackage{algpseudocode}  % for algorithmic pseudocode

\graphicspath{{./images/}}

\title{GraMFedDHAR: Graph Based Multimodal Differentially Private Federated  HAR}

\author{
 Labani~Halder\\
  CVPR Unit\\
  Indian Statistical Institute\\
  Kolkata,  700108 \\
  \texttt{labanihalder1@gmail.com} \\
   \And
 Tanmay Sen \\
  SQC \& OR Unit\\
  Indian Statistical Institute\\
  Kolkata,  700108  \\
  \texttt{sentanmay518@gmail.com} \\
  \And
 Sarbani~Palit \\
    CVPR Unit\\
  Indian Statistical Institute\\
  Kolkata,  700108 \\
  \texttt{palitsarbani@gmail.com} \\
}

\begin{document}

\maketitle
\begin{abstract}
Human Activity Recognition (HAR) using multimodal sensor data remains challenging due to noisy or incomplete measurements, scarcity of labeled examples, and privacy concerns. Traditional centralized deep learning approaches are often constrained by infrastructure availability, network latency, and data sharing restrictions. While federated learning (FL) addresses privacy by training models locally and sharing only model parameters, it still has to tackle issues arising from the use of  heterogeneous multimodal data and differential privacy requirements. In this article, a Graph-based Multimodal Federated Learning framework, GraMFedDHAR, is proposed for HAR tasks. Diverse sensor streams such as a pressure mat, depth camera, and multiple accelerometers are modeled as modality-specific graphs, processed through residual Graph Convolutional Neural Networks (GCNs), and fused via attention-based weighting rather than simple concatenation. The fused embeddings enable robust activity classification, while differential privacy safeguards data during federated aggregation. Experimental results show that the proposed MultiModalGCN model outperforms the baseline MultiModalFFN, with up to 2\% higher accuracy in non-DP settings in both centralized and federated paradigms. More importantly, significant improvements are observed under differential privacy constraints: MultiModalGCN consistently surpasses MultiModalFFN, with performance gaps ranging from 7–13\% depending on the privacy budget and setting. These results highlight the robustness of graph-based modeling in multimodal learning, where GNNs prove more resilient to the performance degradation introduced by DP noise. 
\end{abstract}

% keywords can be removed

\section{Introduction}
Human Activity Recognition (HAR) is now central to applications in smart homes, fitness tracking, healthcare, and human-computer interaction. However, real-world HAR systems face few persistent challenges: heterogeneous data, privacy preservation, and robust multimodal learning. Practical HAR systems must deal with diverse sensors, missing modalities, and client-specific constraints \cite{das2020mmhar,xiong2022unified,yang2024cross,garain2022differentially}. Centralized ensemble and fusion methods are effective in controlled environments but are often impractical for federated learning. This is because modalities are distributed across clients and cannot be centrally concatenated for training\cite{das2020mmhar}.
% Centralized ensemble and fusion methods, which excel in lab environments, often fail to generalized when modalities are distributed across clients \cite{das2020mmhar}.
% A robust solution needs to align and fuse data from different sources while accommodating missing modalities. 
A robust framework is required to be developed that can combine different data types and work even when some data is missing. Another major challenge is privacy preservation for multi-modal sequential data \cite{abadi2016deep,el2022differential,wei2020federated}. Although federated learning (FL) mitigates privacy risks by training models locally and sharing only model updates, it is still vulnerable to inference attacks and leakage of sensitive activity patterns. Traditional Differential Privacy (DP), designed for i.i.d. data, struggles to account for temporal correlations, often misestimating privacy loss in continuous activity streams \cite{roy2023temporal,el2022differential,abadi2016deep}. Thus, HAR requires federated solutions that are not only multimodal-aware but also enhanced with temporally sensitive privacy guarantees. Multimodal HAR benefits from graph-based modeling, which naturally captures spatial and relational dependencies across different sensor modalities. Graph Convolutional Networks (GCNs) have shown strong potential in capturing contextual structure in HAR tasks \cite{mondal2020new,bandyopadhyay2025mharfedllm,wang2023body,yang2022activity,sarkar2021grafehty}. However, adapting GCNs to FL settings introduces new risks, as sharing graph parameters can leak structural information \cite{ahmad2021graph,sarkar2021grafehty}. Developing communication-efficient and privacy-aware federated GCN protocols is therefore essential.

% An effective solution requires novel temporal-aware privacy accounting and noise mechanisms that respect composition over time while preserving model utility. Using graph-based formulations, naturally capture spatial and relational structures in HAR and improve contextual inference \cite{mondal2020new,bandyopadhyay2025mharfedllm,wang2023body,yang2022activity,sarkar2021grafehty} in federated environment. 
% ``Federatedizing'' Graph Convolutional Neural Networks (GCNs) risks leaking structural information during communication of graph parameters \cite{mondal2020new,ahmad2021graph,sarkar2021grafehty}. 
% A viable solution requires communication-efficient, privacy-aware federated GCN protocols, potentially incorporating structural-DP or edge-level obfuscation to mitigate information leakage. % Existing HAR datasets and benchmarks also lack realism. They often fail to include key real-world factors like non-IID label distributions, systematic modality dropouts, asynchronous sampling, and device-level constraints such as energy and latency \cite{kulsoom2022review, ek2022lightweight,bandyopadhyay2025mharfedllm,zakariyya2025differentially}.

Existing HAR datasets and benchmarks lack realism because they often fail to include key real-world factors such as non-IID data distributions, systematic data loss from certain sensors, unsynchronized data collection, and device-level constraints like limited energy and latency  \cite{kulsoom2022review, ek2022lightweight,bandyopadhyay2025mharfedllm,zakariyya2025differentially}. HAR systems also face additional concerns including high labeling costs, vulnerability to adversarial attacks, and the need for explainability and fairness in sensitive domains such as healthcare \cite{sarkar2021grafehty,wei2020federated,wang2023body,garain2022differentially}. These broader challenges highlight the importance of developing multimodal HAR frameworks that preserve privacy, are robust, and are trustworthy.

% This makes it difficult to fairly evaluate the on-device viability and energy trade-offs of proposed solution \cite{ek2022lightweight,bandyopadhyay2025mharfedllm,zakariyya2025differentially}. The high cost of data labeling is a significant barrier. While centralized self-supervised learning can help \cite{abdi2024unsupervised,yang2015deep,singh2020deep,vaswani2017attention}, applying it in a federated setting across heterogeneous modalities is an emerging field with new privacy challenges. Additionally, HAR systems are vulnerable to adversarial threats like sensor spoofing and malicious clients \cite{sarkar2021grafehty,wei2020federated}. Few studies have combined robust defenses against these threats with multimodal privacy guarantees. Finally, ensuring explainability and fairness is critical requirements, especially in healthcare, but can be complicated by graph-based and multimodal models \cite{wang2023body,garain2022differentially}.

In this article, we propose GraMFedDHAR, a Graph-based Multimodal Federated Learning framework for differentially private HAR. In our framework, each sensor stream (e.g., pressure mats, depth cameras, accelerometers) is represented as a modality-specific graph and processed through residual GCNs. Their embeddings are fused via attention-based weighting, enabling robust multimodal activity classification. To protect sensitive information, DP is integrated into the FL aggregation process, ensuring user-level protection. Extensive experiments demonstrate that graph-based multimodal modeling is significantly more resilient than feedforward baselines under DP constraints. The major contributions of this work are summarized as follows:
\begin{itemize}
    \item A novel hybrid graph construction strategy for multimodal HAR is proposed. This approach models temporal dependencies by linking consecutive time windows, while spatial dependencies across sensor modalities are captured through a distance-based adjacency formulation.  
    \item DP is integrated into the FL process. This is achieved by perturbing client-side updates before aggregation, which provides formal user-level privacy guarantees while preserving recognition performance. 
    \item It may also be emphasized that this work is the first to perform the integration of DP and FL on the multimodal graph based HAR task.
\end{itemize}

% In this work, the major contributions made can be summarized as:
% \begin{itemize}
%     \item A novel hybrid graph construction strategy for multimodal HAR is proposed. This approach models temporal dependencies by linking consecutive time windows, while spatial dependencies across sensor modalities are captured through a distance-based adjacency formulation.  
%     \item Differential privacy is integrated into the federated learning process. This is achieved by perturbing client-side updates before aggregation, which provides formal user-level privacy guarantees while preserving recognition performance. 
% \end{itemize}

% It may also be emphasized that this work is the first to perform the integration of DP and federated learning on the mutlimodal graph based HAR task.

\section{Related Works}  In the existing literature, the problem of HAR is typically categorized as a traditional machine learning and deep learning based supervised or unsupervised classification problem. Traditional machine learning approaches using smart sensors and wearable devices for healthcare monitoring, leverage body motion and vital sign data to recognize daily activities. However, performance is often limited by the variability of the data set and the lack of a universally optimal algorithm \cite{subasi2020human}. A comprehensive review on HAR  is provided in \cite{kulsoom2022review}. Apart from highlighting the challenges of accurately detecting complex activities, it further analyzes vision-based and non-vision-based sensors, along with various traditional machine learning techniques applied to different HAR applications. 
% The authors in \cite{abdi2024unsupervised} used Latent Dirichlet Allocation (LDA), an unsupervised statistical method, to convert activity signals into a sequence of discrete symbols through vector quantization. 
 \cite{singh2020deep} proposes Deep-ConvLSTM that learns time-dependent sequence context using a self-attention mechanism for the HAR problem, with applications in healthcare. Recently, Transformer-based models \cite{vaswani2017attention} have gained significant traction for addressing time series problems like HAR. \cite{ek2022lightweight} proposed the Human Activity Recognition Transformer (HART), a lightweight, sensor-wise Transformer optimized for IMU-based HAR on mobile devices, achieving higher accuracy than state-of-the-art methods. Mondal et al. \cite{mondal2020new} proposed a transductive graph-based framework for the HAR task by converting multi-variate time series sensor data into a graph. The model's robustness is also verified on imbalanced data. A comprehensive survey on skeleton-based action recognition using graph convolutional neural networks for video data is provided in \cite{ahmad2021graph}. \cite{wang2023body} introduces body RFID skeleton-based activity recognition using GCN, where activities are modeled as RFID skeleton activity graphs achieving superior recognition performance compared to existing HAR methods.  \cite{yang2022activity} proposes a deep CNN to automatically extract discriminative features from optimal activity graphs for HAR, effectively capturing correlations between activities. 

Multi-modal activity recognition has gained significant attention for its ability to leverage complementary information and improve robustness to missing or corrupted data. \cite{das2020mmhar} proposes MMHAR-EnsemNet, a multi-modal HAR framework that integrates skeleton, RGB, and inertial signals using CNN and LSTM-based networks, with outputs combined in an ensemble achieving superior performance for  experiments on UTD-MHAD and Berkeley-MHAD datasets. The authors in \cite{xiong2022unified}  propose a unified FL framework with co-attention–based fusion and Model-Agnostic Meta-Learning (MAML) driven personalization, demonstrating superior performance on multimodal activity recognition tasks. A federated learning (FL) framework MHARFedLLM for multi-modal HAR using a federated large language model by introducing a novel attention based fusion architecture was proposed by \cite{bandyopadhyay2025mharfedllm}.  
 % The authors in \cite{wang2025privacy} proposed a privacy-preserving FL framework that enables healthcare institutions to train diagnostic models collaboratively without sharing raw data, while effectively fusing heterogeneous medical modalities.  
 Cross-modal federated HAR (FHAR) is addressed by \cite{yang2024cross}, who propose the Modality-Collaborative Activity Recognition Network (MCARN) that learns both global and modality-specific activity classifiers. Differential privacy (DP) \cite{roy2023temporal,  zakariyya2025differentially} mitigates the risk of adversarial leakage in federated learning with formal privacy guarantees. \cite{garain2022differentially} proposes a DP based HAR framework on smartphone accelerometer data using a deep Multi Layer Perceptron (DMLP) algorithm, achieving competitive accuracy while ensuring strong privacy guarantees. Sarkar et al. \cite{sarkar2021grafehty} proposed GraFeHTy, a framework that incorporates federated graph learning to address noisy, low-data challenges and privacy concerns in HAR. However no work has yet explored federated differential privacy for graph based multimodal human activity recognition.

The paper is organized as follows: Section~II presents the related works. Section~III describes the preliminary concepts while the  problem statement and proposed work are discussed in Section~IV and~V, respectively. Section~VI covers the simulation and evaluation. Section VII concludes the article.

\section{Preliminary}
This Section briefly discusses the Graph Neural Network, Multimodal Federated Graph Learning and Differential privacy.
\subsection{\textbf{Graph Neural Network}}To capture the complex interdependencies among data points, real-world problems are often modeled using graphs where each instance is treated as a node connected to others by an edges based on similarities measures. A graph is a tuple $\mathcal{G} = (\mathcal{V,E,X})$, where $\mathcal{V}$ is the node set, $\mathcal{E}$- edge set, and $\mathcal{X}\in \mathbb{R}^{N\times d}$- the feature matrix ($N$ nodes, $d$ features per node). Each node $i$ has an initial embedding $h^0_{i} = \mathcal{X}_{i}$. After the $k^{th}$ iteration (or $k$ layer), the embedded representation of the node $x_{i}$ is updated as follows \cite{kipf2016semi} $
    h_{i}^{(k+1)} = \sigma \left( W_{(k)}\sum_{u \in \mathcal{N}(i)} \frac{h_{i}^{(k)}}{|\mathcal{N}(i)|} +{B_{(k)}h_{i}^{(k)}}\right)$, where $h_{i}^{k+1}$ is the hidden state of the node $x_{i}$ at layer $k$, $\mathcal{N}(i)$ denotes the set of neighboring nodes of $x_{i}$, $W_{k}$ and $B_{k}$ are the learnable weight matrices, $\sigma(\cdot)$ is a non-linear activation. The compact matrix form is $H^{(k+1)} = \sigma\left( \tilde{D}^{-\frac{1}{2}} \tilde{A} \tilde{D}^{-\frac{1}{2}} H^{(k)}W^{(k)} +B^{(k)}H^{(k)} \right)
    $ with $H^{0} = \mathcal{X}$. In this work, node-level classification is performed.
\subsection{Multi-modal Federated Graph Learning} The sensor signals are multimodal time series in nature:
$X = \{X^{(1)}, X^{(2)}, \ldots, X^{(M)}\}, \quad X^{(m)} \in \mathbb{R}^{T \times d_m}$, where $m = 1,\cdots, M$ is the number of modalities, $T$ time steps and $d_m$ the dimension of each modalities. In federated learning, each client $c$ with dataset $\mathcal{D}_{c}$ trains locally. The global training objectives is $
    \min_{\theta} \sum_{c = 1}^{N} p_c \mathcal{L}(\mathcal{D}_c; \theta), \quad p_c = \frac{|\mathcal{D}_c|}{\sum_{j\in N}  |\mathcal{D}_j|}$. The central sever aggregates update $\theta$, avoiding access to any of client's raw data. Each client transforms its graph $\mathcal{G}_{c} =\left(\mathcal{V}_{c},\mathcal{E}_{c}\right)$, where nodes $\mathcal{V}_{c}$ represent time windows and $\mathcal{E}_{c}$ defined by the distance-based similarity. Each node $x_{i}$ is a multi-modal feature vector 
$\mathbf{x}_{i} = \text{Fuse}_M\left( \phi^{(1)}(x_v^{(1)}), \ldots, \phi^{(M)}(x_v^{(M)}) \right)$, where $\phi^{(m)}$ are modality-specific encoders, and $\text{Fuse}_M$ is a attention based fusion mechanism. This multi-modal feature vector is then used to classify the activity nodes.

\subsection{Differential Privacy} A randomized mechanism $\mathcal{M}$ \cite{el2022differential, wei2020federated, abadi2016deep} is $(\varepsilon, \delta)$-differentially private if two neighboring dataset $\mathcal{D}, \mathcal{D}'$ and any event $\mathcal{S}$: $\Pr[\mathcal{M}(\mathcal{D}) \in S] \leq e^{\varepsilon} \Pr[\mathcal{M}(\mathcal{D}') \in S] + \delta$. 
Rényi Divergence of order $\alpha$ is adopted in this work, defined by $\mathcal{D}_\alpha(M \| N) = \frac{1}{\alpha - 1} \log \mathbb{E}_{x \sim N} \left[ \left( \frac{M(x)}{N(x)} \right)^\alpha \right]$ with $\alpha>1$. If $\mathcal{M}:\mathcal{X}\rightarrow \mathbb{R}$ is said to be $(\alpha,\gamma)-RDP$, then under T-fold composition it implies $(\varepsilon, \delta)$-DP, where
$\mathcal{D}_\alpha(\mathcal{M}(\mathcal{\mathcal{D}})||\mathcal{M}(\mathcal{\mathcal{D'}}) \leq\gamma$. This ensures each client's contribution is bounded and private.
\section{Problem Statement} 
\textbf{Problem Definition:} Multi-modal time series inputs, which are frequently privacy-sensitive and dispersed across user devices, are used in HAR utilizing wearable sensor data. Centralized approaches risk data leakage and scalability issues. A privacy-conscious option is provided by FL; however, there are still issues with $1)$ handling heterogeneous data $2)$ identifying temporal and structural relationships, and $3)$ mitigating gradient-based privacy threats. This work GraMFedHAR addresses these challenges for secure and efficient HAR.\\

% \textbf{Threat Model:} Herein, a semi-honest(honest-but-curious) adversary model is adopted. The central server correctly follows FL protocol while attempting to obtain private client's data from the model updates that are delivered. Clients are assumed to be honest and do not conspire. To mitigate such risk, DP is attached during local model updates. 

\noindent \textbf{Design Goals:} GraMFedHAR is a novel framework designed to achieve the following goals in HAR application $1)$ multimodal graph learning, and $2)$ differential privacy for gradient protection.
\section{Proposed Scheme}
In this work, GraMFedHAR, a multimodal graph-based federated learning (FL) framework with client-level differential privacy (DP) is proposed. The framework integrates modality-specific graph construction, attention-based fusion, and DP-preserving updates under the federated setting.
\subsection{System Overview}
Each client $C_c$ maintains multimodal streams $\{X_c^m\}_{m=1}^M$ with activity labels $Y_c^m$. Data is segmented into windows, transformed into embeddings, and fused into node features $H_c$. A client-specific graph $\mathcal{G}_c=(\mathcal{V}_c,\mathcal{E}_c,H_c)$ is constructed using time window as node and distance-based similarity links. Clients train GCNs locally with graph-regularized loss and add Gaussian noise to clipped updates for DP. The server aggregates these updates using FedAvg.
\subsection{Algorithmic Workflow}
The training protocol is presented in Algorithm~\ref{algo:gramfedp}, which consists of three steps: (i) graph-based feature propagation via GCN layers, (ii) local DP training, and (iii) secure aggregation.

\begin{algorithm}[hbt]
\caption{\textsc{GraMFedDHAR: Federated Multimodal GCN with Differential Privacy}}
\label{algo:gramfedp}
\textbf{Input:} Client set $\mathcal{C}$; total rounds $L$; client sampling fraction $q$; DP parameters $(C,\sigma)$; learning rate $\eta$, $B$ batch size.\\
\textbf{Output:} Global model $w_{\text{global}}$.
\begin{algorithmic}[1]
\State Initialize global weights $w_{\text{global}}$.
\For{$\ell = 1$ to $L$} 
    \State Randomly sample subset $\mathcal{S}_\ell \subset \mathcal{C}$ with $|\mathcal{S}_\ell| = \lfloor q|\mathcal{C}|\rfloor$.
    \For{each client $c \in \mathcal{S}_\ell$ in parallel} 
        \State \textbf{Graph Encoding:} Build local graph $\mathcal{G}_c$ from multimodal input $X_c^{1:M}$ and propagate features using GCN layers.
        \State \textbf{Local Training with DP:} 
        \begin{enumerate}
            \item Compute cross-entropy loss:
            $
            \mathcal{L}_c = \text{CE}(Y_c,\hat{Y}_c).
            $
            \item Compute local gradient $g_c = \nabla_w \mathcal{L}_c(w_{\text{global}})$.
            \item Clip gradient: $g_c \gets g_c / \max(1, \|g_c\|_2 / C)$.
            \item Add Gaussian noise: $\tilde{g}_c = g_c + \mathcal{N}(0, \sigma^2 C^2 I)$.
            \item Update local weights: $w_c \gets w_{\text{global}} - \eta \tilde{g}_c$.
        \end{enumerate}
        \State Send update $\Delta w_c = w_c - w_{\text{global}}$ to server.
    \EndFor
    \State \textbf{Aggregation:} 
    \(
    w_{\text{global}} \gets w_{\text{global}} + \tfrac{1}{|\mathcal{S}_\ell|} \sum_{c \in \mathcal{S}_\ell} \Delta w_c
    \)
    % \State \Return $w_{\text{global}}$
\EndFor \\
\Return $w_{\text{global}}$
% % \State \Return $w_{\text{global}}$.
% \State \Return $w_{\text{global}}$ \\
\end{algorithmic}
\end{algorithm}

\subsection{Global Objective and Assumptions}
% The global objective is defined as $
% F(w) = \frac{1}{N}\sum_{c=1}^{N}\mathbb{E}\big[\mathcal{L}_c(w;\mathcal{G}_c)\big]$, where the local client loss is given by:
% $\mathcal{L}_c = \mathrm{CE}(Y_c,\hat{Y}_c) + \lambda \sum_{(i,j)\in\mathcal{E}_c}\|H_i - H_j\|^2$,
% with $\mathrm{CE}(\cdot)$ denoting the cross-entropy loss between predicted and true activity labels, and the second term enforcing \emph{graph smoothness} by penalizing feature differences along edges in $\mathcal{G}_c$.
The global objective is defined as 
\[
F(w) = \frac{1}{N}\sum_{c=1}^{N}\mathbb{E}\big[\mathcal{L}_c(w;\mathcal{G}_c)\big],
\]
where the local client loss is given by
\[
\mathcal{L}_c = \mathrm{CE}(Y_c,\hat{Y}_c),
\]
with $\mathrm{CE}(\cdot)$ denoting the cross-entropy loss between predicted and true activity labels. 
The optimal model parameters are denoted by
\[
w^\star = \arg\min_w F(w).
\]
Here, the graph structure $\mathcal{G}_c$ influences the training implicitly through GCN message passing layers, rather than through an explicit smoothness regularizer.

% \noindent\textbf{Assumptions:} There is some assumption $1)$ $F$ is $L$-smooth: $\|\nabla F(x)-\nabla F(y)\|\le L\|x-y\|$, $2)$$F$ is $\mu$-strongly convex,$3)$ Gradients are bounded: $\|\nabla \mathcal{L}_c(w)\| \le G$, and $4)$ Local updates are clipped at $C$ before adding Gaussian noise for DP.
% \subsection{Convergence Guarantee}
% Let $m=\lfloor qN\rfloor$ denote the number of sampled clients per communication round, each client train a GCN with graph-regularized loss and Gaussian noise added to clipped updates for differential privacy. With a step size $\eta$, Algorithm~\ref{algo:gramfedp} achieves linear convergence up to a DP-induced noise \cite{convergence}, $\mathbb{E}\big[\|w_\ell - w^\star\|^2\big] \le \Big(1 - \tfrac{\eta\mu}{2}\Big)^\ell \|w_0 - w^\star\|^2 + \frac{4}{\mu^2}\Bigg(\underbrace{\frac{\sigma_g^2}{mB}}_{\text{stochasticity}} + \underbrace{\frac{\zeta^2}{m}}_{\text{heterogeneity}} + \underbrace{\frac{d\sigma^2C^2}{m}}_{\text{DP noise}}\Bigg)
% $, where $\sigma_g^2$ captures the variance from mini-batch stochastic gradients, $\zeta^2$ reflects client drift due to non-IID multimodal data, and  $d\sigma^2C^2/m$ corresponds to the DP-induced error floor from Gaussian noise added to clipped updates.
\noindent\textbf{Assumptions:} We make the following standard assumptions:
\begin{enumerate}
    \item $F$ is $L$-smooth: $\|\nabla F(x)-\nabla F(y)\|\le L\|x-y\|$,
    \item $F$ is $\mu$-strongly convex,
    \item Gradients are bounded: $\|\nabla \mathcal{L}_c(w)\| \le G$,
    \item Local updates are clipped at $C$ before adding Gaussian noise for DP.
\end{enumerate}

\subsection{Convergence Guarantee}
Let $m=\lfloor qN\rfloor$ denote the number of sampled clients per communication round. 
Each client trains a GCN locally with cross-entropy loss and adds Gaussian noise to clipped updates for differential privacy. 
With step size $\eta$, Algorithm~\ref{algo:gramfedp} achieves linear convergence up to a DP-induced error floor~\cite{convergence}:
$$
\mathbb{E}\big[\|w_\ell - w^\star\|^2\big] \le 
\Big(1 - \tfrac{\eta\mu}{2}\Big)^\ell \|w_0 - w^\star\|^2 
+ \frac{4}{\mu^2}\left(
\underbrace{\tfrac{\sigma_g^2}{mB}}_{\text{stochasticity}}
+ \underbrace{\tfrac{\zeta^2}{m}}_{\text{heterogeneity}}
+ \underbrace{\tfrac{d\sigma^2C^2}{m}}_{\text{DP noise}}
\right),
$$

where $\sigma_g^2$ captures the variance from mini-batch stochastic gradients, $\zeta^2$ reflects client drift due to non-IID multimodal data, and $d\sigma^2C^2/m$ corresponds to the DP-induced error floor from Gaussian noise added to clipped updates.

\section{Simulation and Evaluation}
\subsection{Implementation Details and Dataset Preparation} 
\subsubsection{Dataset} To evaluate the effectiveness of the proposed framework, experiments are conducted on the publicly available MEx dataset, which captures multimodal data from $30$ participants performing $7$ physiotherapy exercises, each lasting up to $60$ seconds. The dataset includes four sensor modalities: two tri-axial accelerometers (placed on the wrist \textit{acw} and thigh \textit{act} sampling at approximately $100$Hz for fine-grained motion data; \textit{depth camera} (\textit{dc}) recordings sampled at around $15$Hz and it is downsampled from $240\times320$ to a $12\times16$ resolution using OpenCV for low-resolution spatial posture tracking; and a \textit{pressure mat} (\textit{pm}) capturing $32\times16$ spatial pressure distributions at ~$15$ Hz to monitor contact patterns. This rich modality data supports robust spatiotemporal modeling for physiotherapy movement analysis. 
\subsubsection{Data Preprocessing and Segmentation} Sensor timestamps were standardized to Unix epoch format with microsecond precision for all sensor modalities. Sampling rates were normalized using linear interpolation: \textit{act, acw} at $100$Hz, and \textit{dc} and \textit{pm} data were resampled to $15$ Hz. Z-score normalization was applied independently to each modality, and timestamps were discard after alignment. Time series were segmented using a $5$-second sliding window with a $2$-second stride ($60\%$ overlap), yielding $500$ frames per widow for act, acw and $75$ for dc and pm. dc and pm features were extracted using a fully connected autoencoder (layers: $512-256-64$, ReLU activations), with the $64$-dimensional latent vector used as the final embedding, Discrete Cosine Transformation (DCT) was applied per axis for act and acw data, retaining $60$ coefficients, resulting in a $180$-dimensional feature vector per window.
\subsubsection{Experimental Setup} The experiments were performed using Python 3.12, with Pytorch 2.6.1 serving as the primary deep learning framework. Additional dependencies included scikit-learn 1.5.1 for data pre-processing and evaluation metrics, as well as pandas and Numpy for data manipulation and numerical computation. The experiments were carried out on a desktop with Intel(R) Core(TM) i$7-12700 (2.10$ GHz) CPU with $16$ GB RAM running Windows-11. 
\subsection{Model Architecture}
\subsubsection{Model Specification} In this work, a Multi-modal Graph Convolutional Network (MultiModalGCN) is used to effectively integrate multiple graph-structured modalities. Each modality is independently passes through a dedicated GCN layer that transforms node features into a shared hidden space, followed by normalization, Relu activation, and dropout. An attention-based fusion mechanism is used to compute per-node attention weights for each modality, allowing the most informative signals to emphasized before fusion into unified node embedding. Nodes are each timestamp window and edges are created between the nodes based on the similarity distance function using threshold based tuning. The model is configured with modality-specific input sizes, a fixed hidden dimension $64$, and a dropout rate of $0.5$.
\subsubsection{Baseline Model} As a baseline, a Multimodal Feedforward Neural Network (MultiModalFFN) is designed without utilizing graph convolutional using same configuration as the MultiModalGCN.  
\subsubsection{Training and Learning algorithms} For both Centralized and Federated learning settings, local training was performed on each client using supervised learning approach with cross-entropy loss and Adam optimizer at a learning rate of $0.01$. During training, forward passes are performed on graph-structured data to generate predictions, and the cross-entropy is backpropagated. In scenarios where Differential Privacy (DP) is enabled, a differentially private stochastic gradient decent (DP-SGD) mechanism was adopted. Specifically, gradients are clipped to a maximum norm and Gaussian noise sampled is added to each gradient. This ensures individual data contributions remain private. After local training, we evaluated the model performance on a held-out test dataset. Federated Learning, each client's local dataset is split into $70\%$ training and $30\%$ testing. After training, model updates are sent to a central server, where they are averaged using FedAvg on update the global model. Centralized Learning, the dataset is split into $70\%$ for training and $30\%$ for testing and $500$ local epochs are used to maintain the same DP configuration in Federated Learning. The model hyper-parameters are described in Table \ref{tab:parameter}.
\begin{table}[h]
\centering
\caption{Hyper-parameters for Federated and Centralized Learning}
\begin{tabular}{lcc}
\toprule
\textbf{Hyper-parameter} & \textbf{Federated} & \textbf{Centralized} \\
\midrule
Learning rate            & 0.01              & 0.001 \\
Number of layers         & 2                 & 2 \\
Hidden units per layer   & 64                & 128 \\
Dropout rate             & 0.5               & 0.5 \\
Batch size               & 32                & 32 \\
Local epochs per client  & 20                & --- \\
Total rounds             & 50                & --- \\
Gradient clipping        & 1                 & 1 \\
Sampling rate for DP     & 0.01              & 0.01 \\
Delta for DP             & $1 \times 10^{-3}$& $1 \times 10^{-3}$ \\
Number of epochs (centralized)&---&500\\
\bottomrule
\end{tabular}
\label{tab:parameter}
\end{table}
\subsubsection{Evaluative metrics} Model performance in both federated and centralized setups was assessed on a held-out set of set using Accuracy and F1 Score. Accuracy measures overall classification correctness, while F1 Score captures the balance between precision and recall. For modals trained under differential privacy, utility loss was additionally calculated to quantify performance degradation introduced by privacy constraints. Utility loss was defined as  $( 1 - \frac{\text{Accuracy}_{\text{DP}}}{\text{Accuracy}_{\text{No DP}}})$, where $\text{Accuracy}_{\text{DP}}$ and $\text{Accuracy}_{\text{No DP}}$ denote the accuracies of the differentially private and non-private models, respectively. This metric captures the trade-off between privacy preservation and model utility. A lower utility loss indicates that privacy had minimal effect on accuracy, while a higher values reflect greater performance degradation.
\subsection{Result and Discussion}
\subsubsection{Impact of DP on Multimodal GCN and FFN Models }
\begin{table}[h]
\centering
\caption{Performance Comparison of GCN and FFN Models with and without Differential Privacy (DP) in Centralized and Federated Settings}
\vspace{0.1cm}
\begin{tabular}{p{2cm} c c c c c c}
\toprule
\multirow{3}{*}{Model} & \multirow{3}{*}{DP Setting} & \multicolumn{2}{c}{Centralized} & \multicolumn{2}{c}{Federated} \\
\cmidrule(lr){3-4} \cmidrule(lr){5-6}
 &  & Accuracy & F1 & Accuracy & F1 \\
\midrule
MultiModalFFN & $\varepsilon=\infty$ & $0.9650$ & $0.9652$ & $0.9850$ & $0.9850$\\
 & $\varepsilon=0.5$& $0.6708$& $0.6672$ & $0.8126$ & $0.8132$ \\
 & $\varepsilon=1.5$ & $0.7308$&$0.7297$ &$0.8863$&$0.8878$\\
\midrule
MultiModalGCN & $\varepsilon=\infty$ & $0.9878$ & $0.9879$ & $0.9941$ & $0.9943$ \\
 & $\varepsilon=0.5$  & $0.8075$ & $0.8077$ & $0.8917$ & $0.8955$ \\
 &$\varepsilon=1.5$ &$0.9011$& $0.9018$&$0.9562$&$0.9579$\\
\bottomrule
\end{tabular}
\label{tab:gcn_ffn_performance}
\end{table}
Table \ref{tab:gcn_ffn_performance} highlights the performance trade-off induced by differential privacy across centralized and federated training for MultiModalFFN and MultiModalGCN models using all four modalities act, acw, dc, and pm. The DP mechanism was implemented using Gaussian noise \textbf{$\varepsilon = 0.5, \sigma = 1.4216, \delta = 1 \times 10^{-3}$} for strong DP, \textbf{$\varepsilon = 1.5, \sigma = 0.8629, \delta = 1 \times 10^{-3}$} for weak DP settings. In the non-private setting, both models perform well, with the GCN outperforming the FFN, especially in the federated settings. However, when DP is applied, a clear performance drop is observed most severe in centralized training. For example, under strong DP ($\varepsilon = 0.5$), centralized GCN accuracy drops by over $18\%$, and FFN by over $29\%$. In contrast, the federated setting demonstrates subtantial resilience on DP-introduced performance degradation. With strong DP, the federated GCN retains ~$10\%$ loss in accuracy and ~$9.88\%$ loss for F1 Score. This demonstrates that GCN outperforms FFN across all conditions, federated learning is more resilient to privacy constraints, and multi modal inputs enhance model robustness, especially in privacy-preserving scenarios.
\subsubsection{Visualization of Fused Embedding under DP via t-SNE } 
\begin{figure}[h]
\centering
\setlength{\tabcolsep}{4pt} % spacing between columns
\renewcommand{\arraystretch}{1} % spacing between rows
\begin{tabular}{c c c}
% Header Row
& \small Centralized & \small Federated \\
\midrule
% Row 1: MultiModalFFN
\rotatebox{90}{\hspace{0.8em}\textbf{MultiModalFFN}} &
\includegraphics[width=0.45\linewidth]{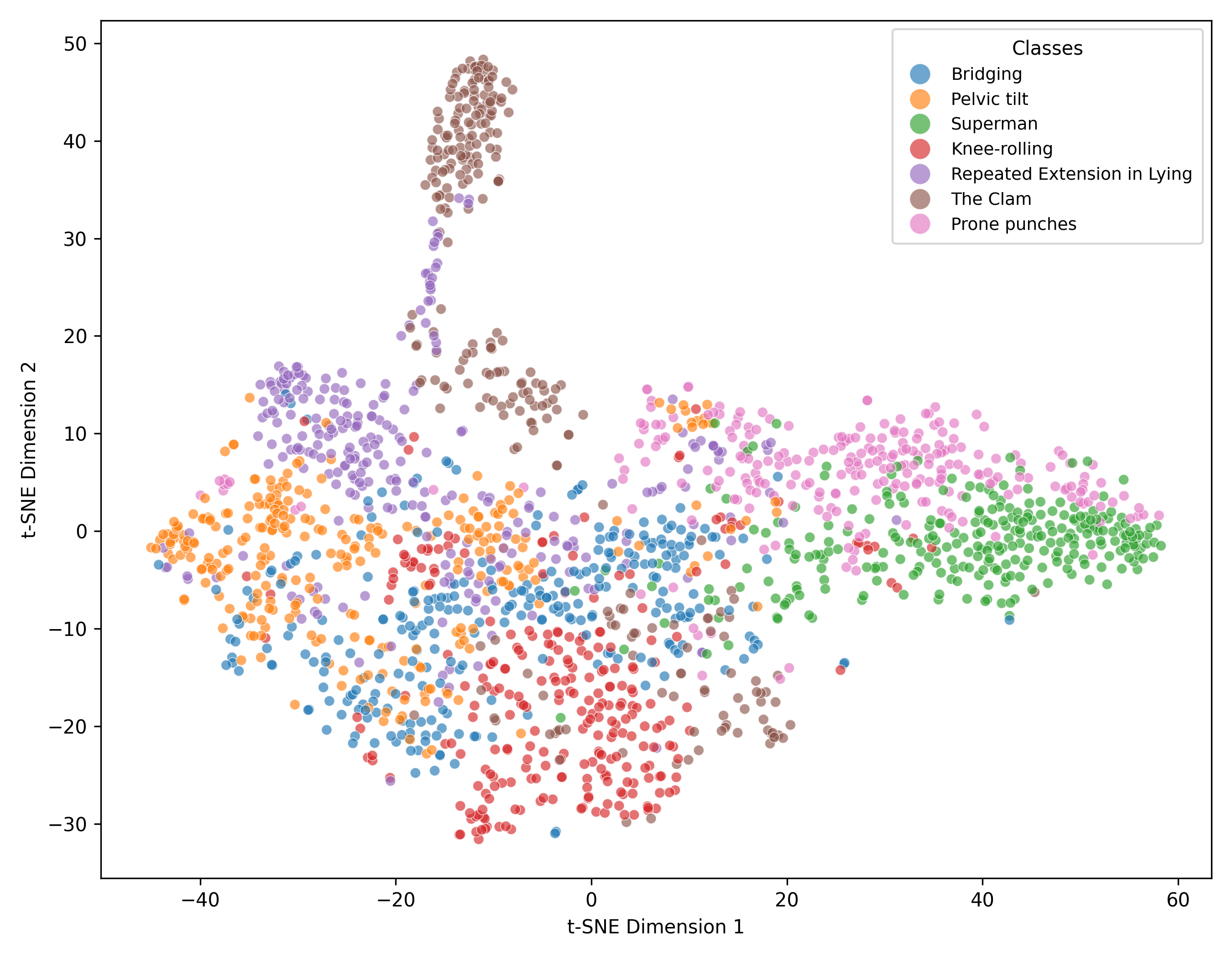} &
\includegraphics[width=0.45\linewidth]{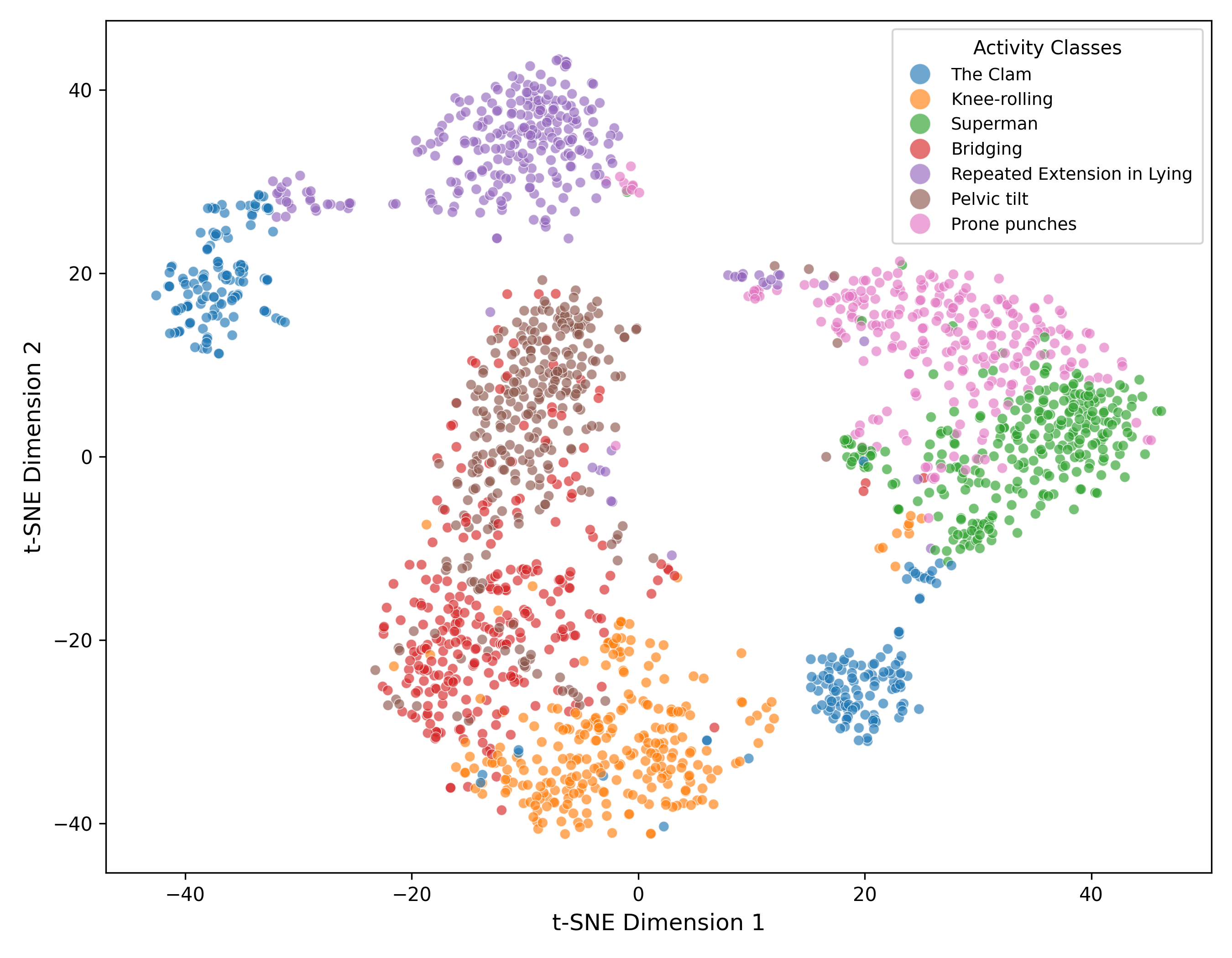}
\\[-0.5em]
% Row 2: MultiModalGCN
\rotatebox{90}{\hspace{0.8em}\textbf{MultiModalGCN}} &
\includegraphics[width=0.45\linewidth]{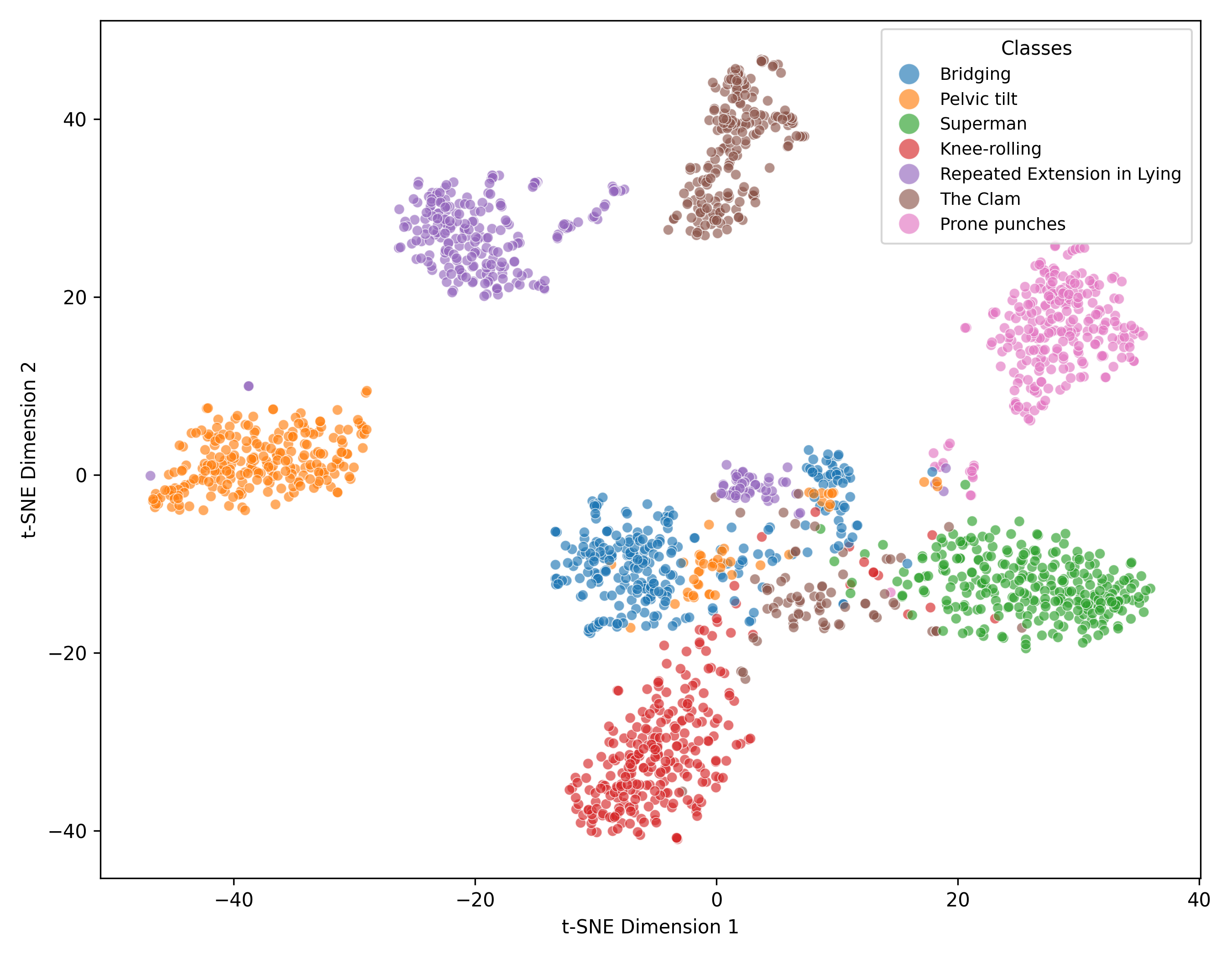} &
\includegraphics[width=0.45\linewidth]{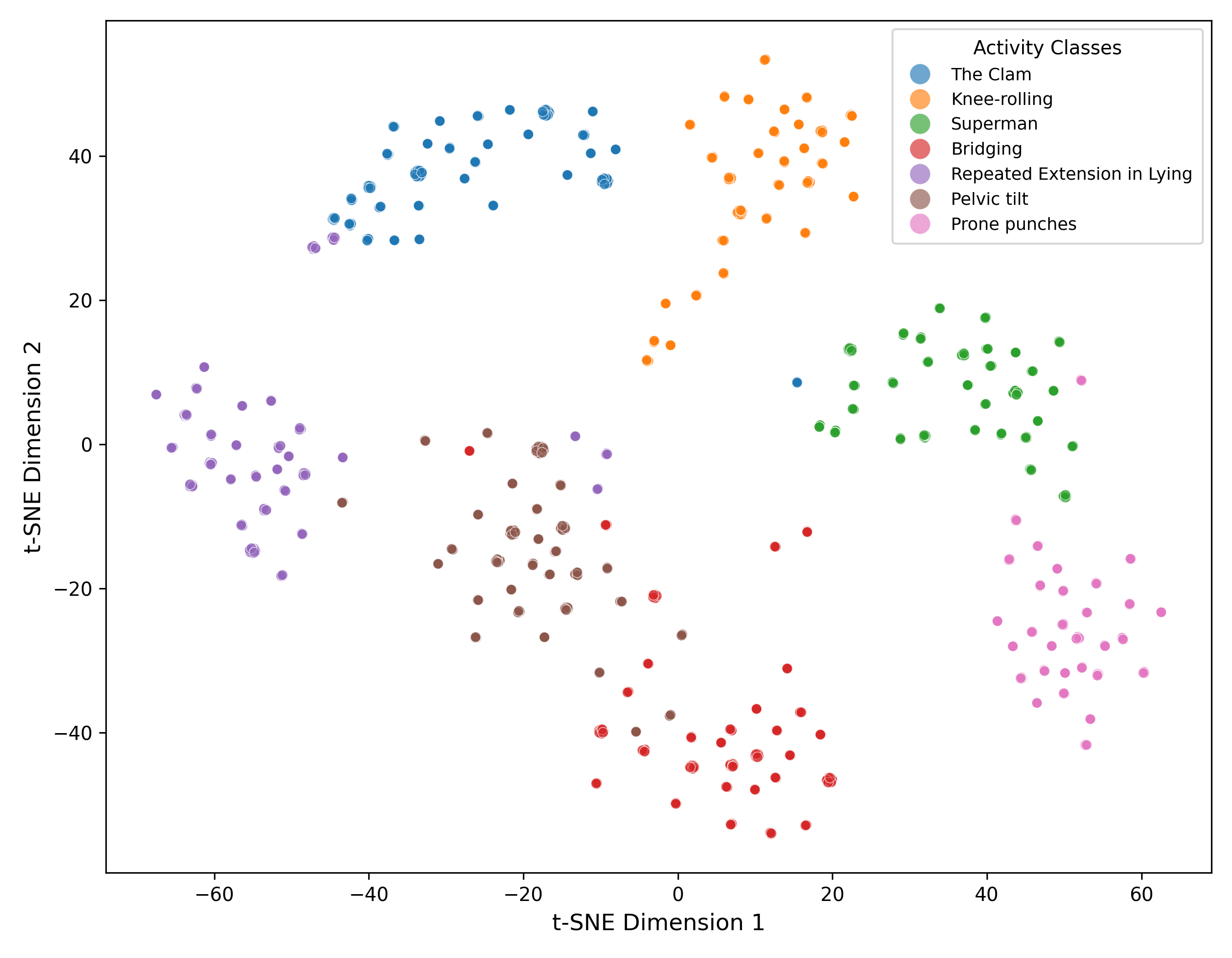}
\end{tabular}
\vspace{0.5em}
\caption{t-SNE visualization under Weak DP ($\varepsilon=2.0, \sigma = 0.7852$) for MultiModalFFN and MultiModalGCN in centralized and federated paradigms.}
\label{fig:tsne_weak_dp1}
\end{figure}
% ==================== Strong DP ====================
\begin{figure}[h]
\centering
\setlength{\tabcolsep}{4pt}
\renewcommand{\arraystretch}{1}
\begin{tabular}{c c c}
% Header Row
& \small Centralized & \small Federated \\
\midrule
% Row 1: MultiModalFFN
\rotatebox{90}{\hspace{0.8em}\textbf{MultiModalFFN}} &
\includegraphics[width=0.45\linewidth]{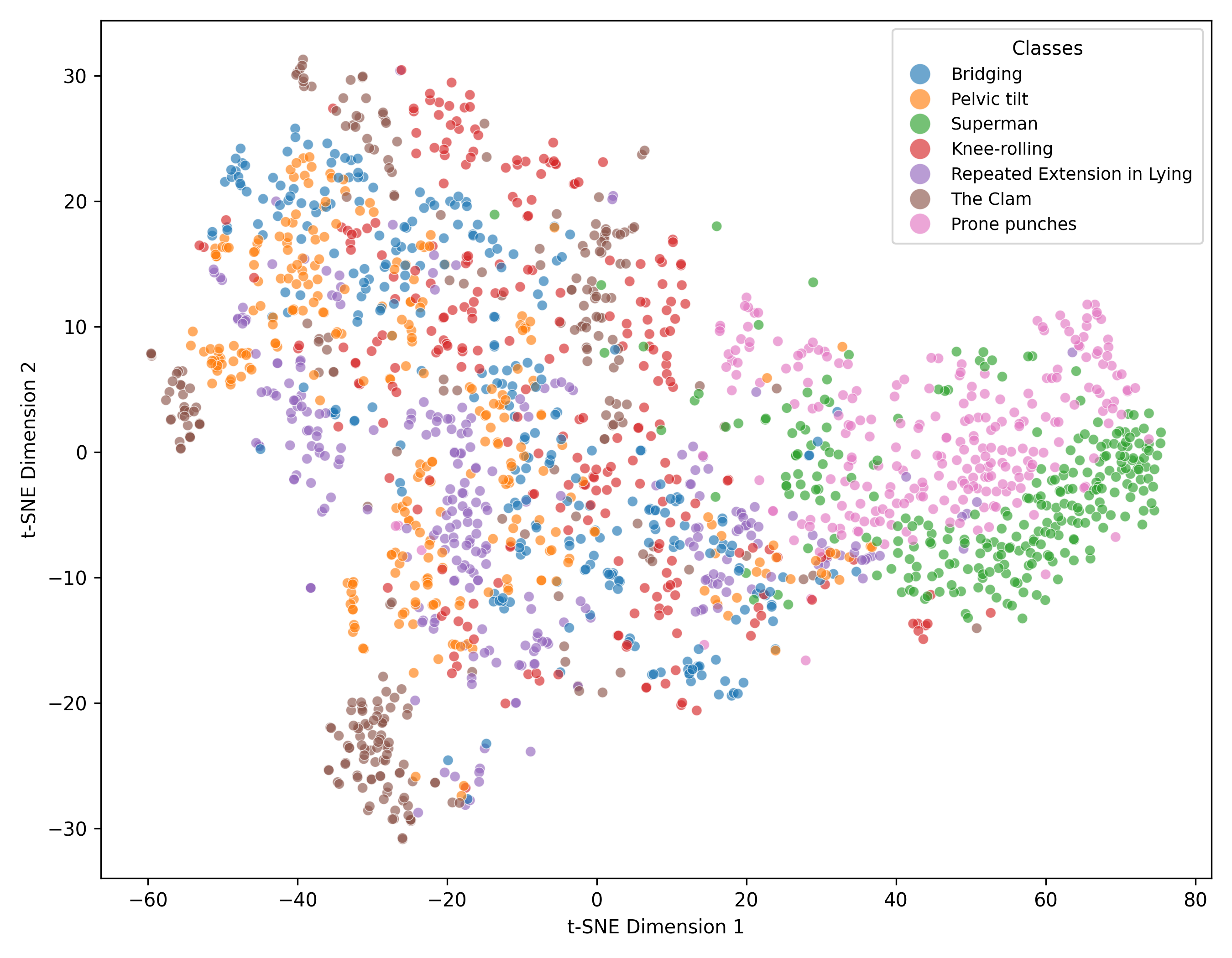} &
\includegraphics[width=0.45\linewidth]{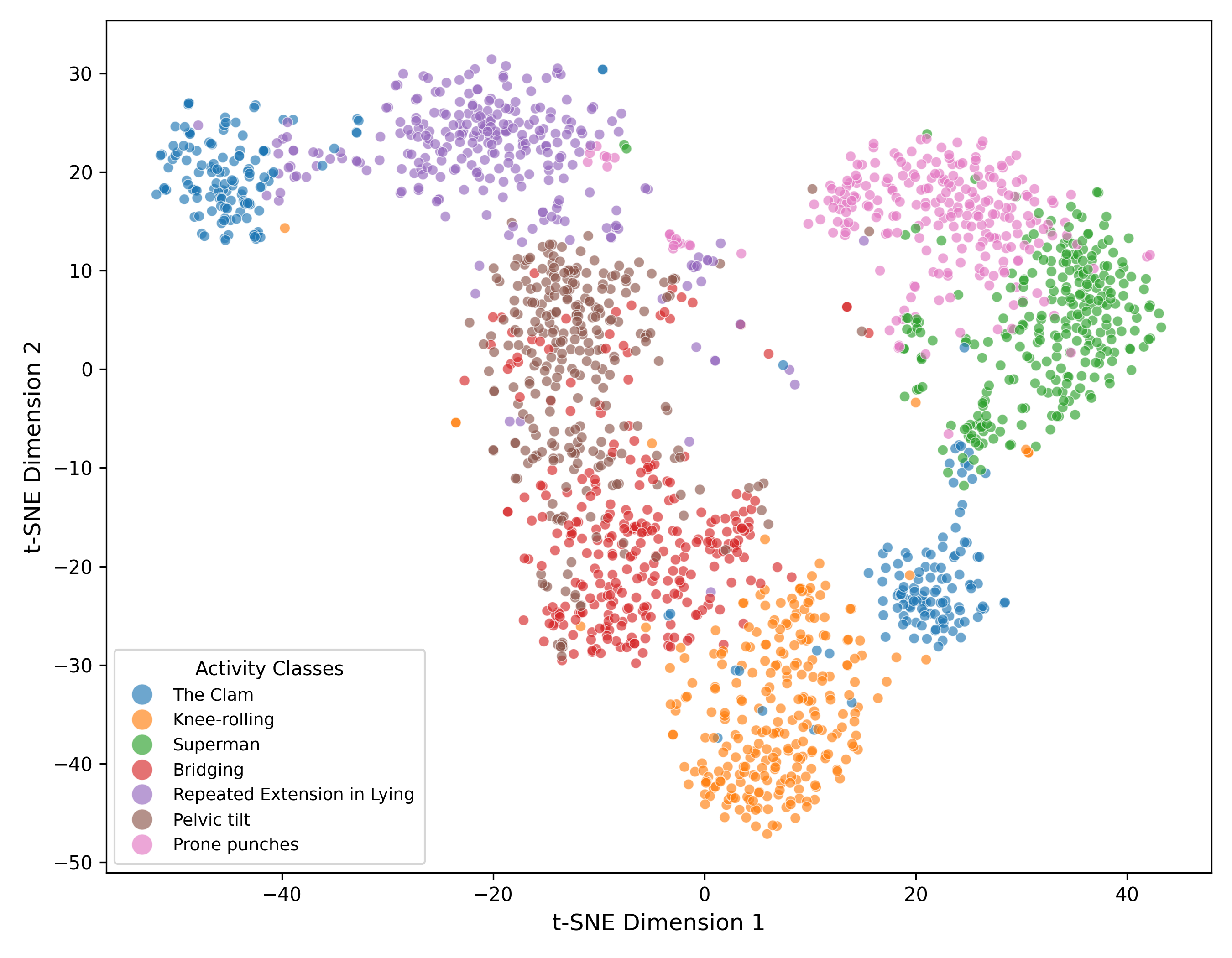}
\\[-0.5em]
% Row 2: MultiModalGCN
\rotatebox{90}{\hspace{0.8em}\textbf{MultiModalGCN}} &
\includegraphics[width=0.45\linewidth]{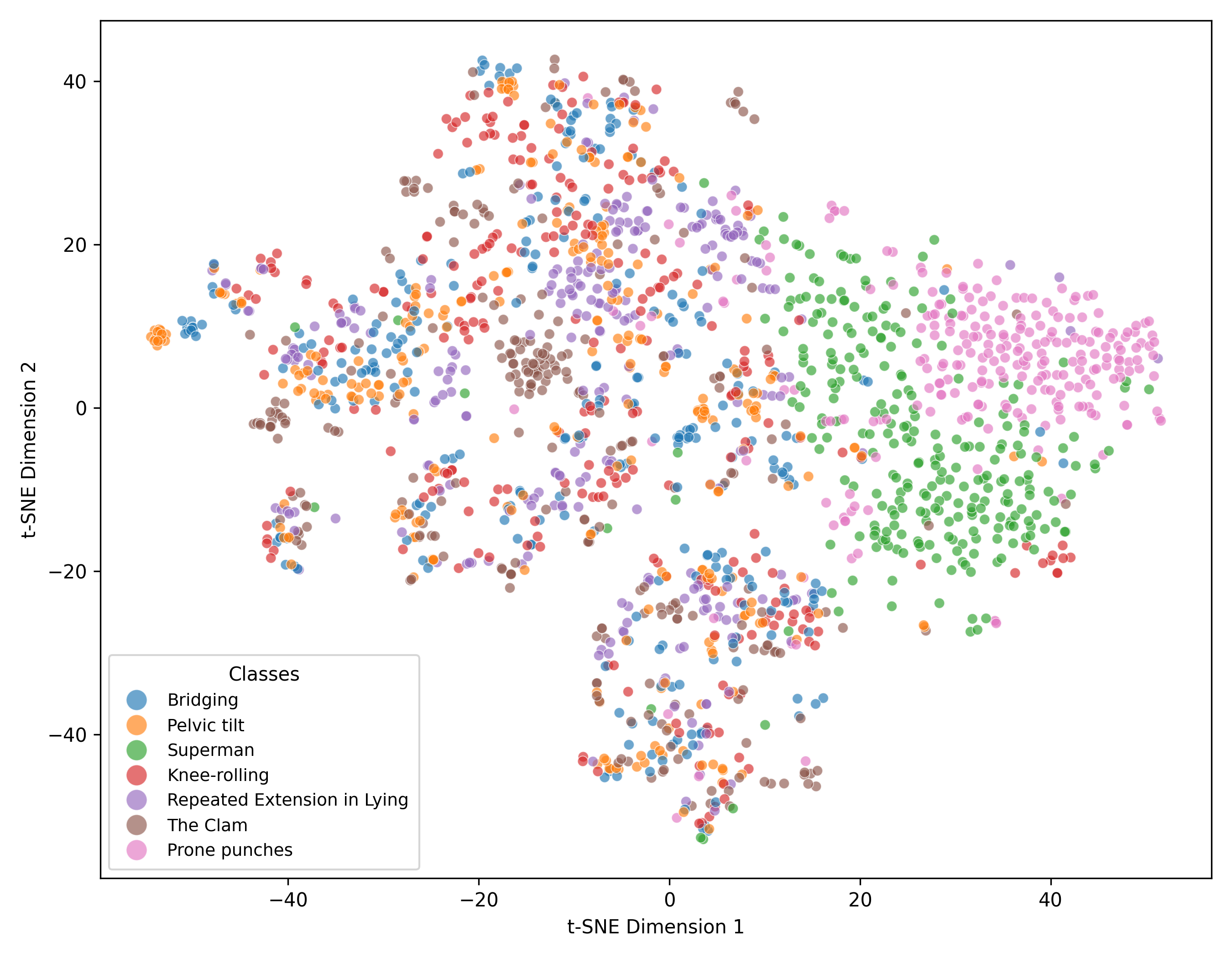} &
\includegraphics[width=0.45\linewidth]{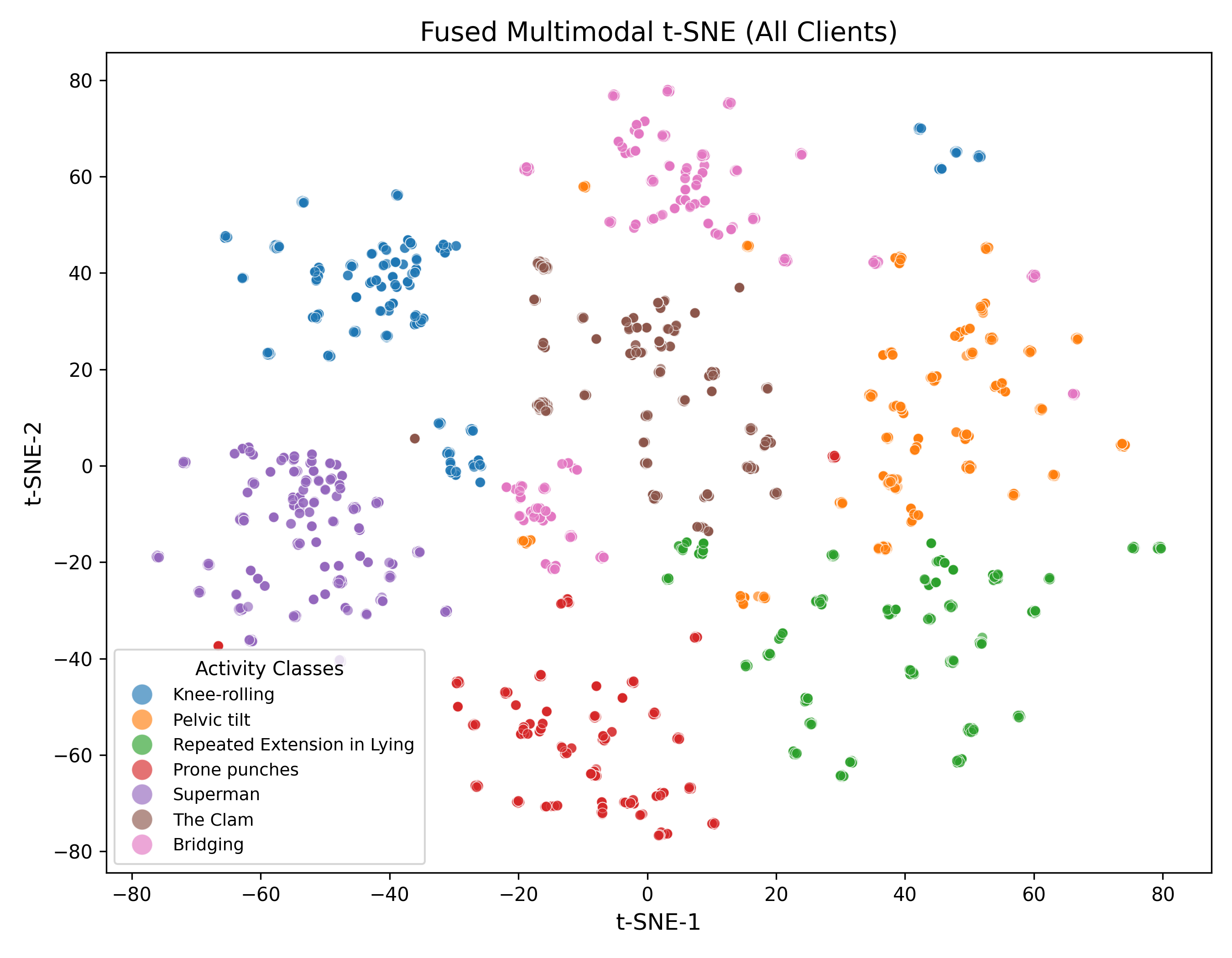}
\end{tabular}
\vspace{0.5em}
\caption{t-SNE visualization under Strong DP ($\varepsilon=0.5, \sigma=1.4216$) for MultiModalFFN and MultiModalGCN in centralized and federated paradigms.}
\label{fig:tsne_strong_dp}
\end{figure}
Figure~(\ref{fig:tsne_weak_dp1}) and (2) present the t-SNE projections of fused multi modal embeddings (act, acw, dc, pm) learned under both weak and strong DP settings. In the centralized setting, both model suffer from noise-induced degradation. The centralized FFN shows significant class overlap and poor clustering, while the centralized GCN offers slightly more structure but remains diffuse, reflecting limited resistance to DP noise. In contrast, the federated setting results in noticeably better embeddings. The federated FFN shows improved class separation, and the federated GCN produces the most distinct and compact clusters, even under strong DP consideration. These results reinforce the quantitative findings: federated learning significantly mitigates the impact of DP noise and also preserves embedding quality and class discriminability, and the integration of graph-based architectures and multimodal inputs further enhances robustness in privacy-preserving scenarios.

\subsubsection{Convergence Under Noise: a comparison of the behaviors under Centralized Learning and Federated Learning} 
% ========== Centralized Learning ==========
\begin{figure}[h]
\centering
\setlength{\tabcolsep}{4pt}
\renewcommand{\arraystretch}{1.2}
\begin{tabular}{c c c}
% Header row
& \small Accuracy & \small Loss \\
\midrule
% Row 1: MultiModalFFN
\rotatebox{90}{\hspace{1em} \textbf{MultiModalFFN}} &
\includegraphics[width=0.45\linewidth]{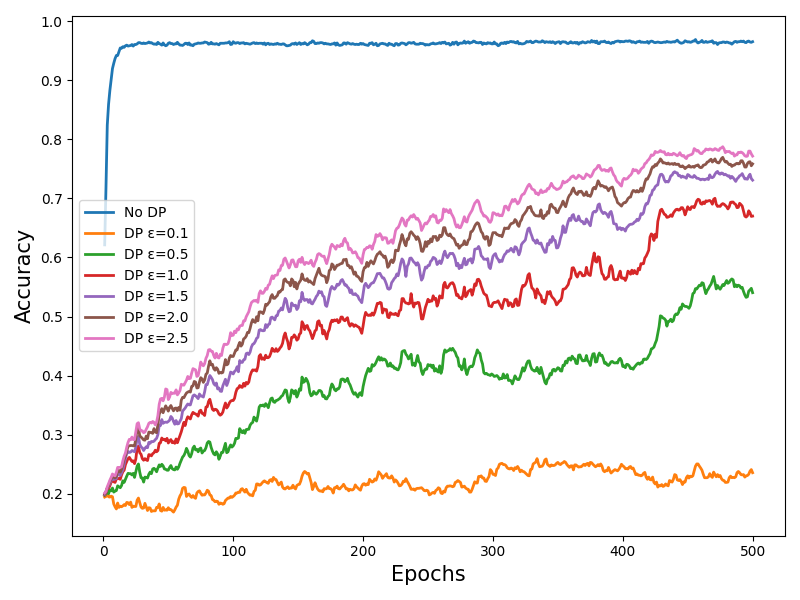} &
\includegraphics[width=0.45\linewidth]{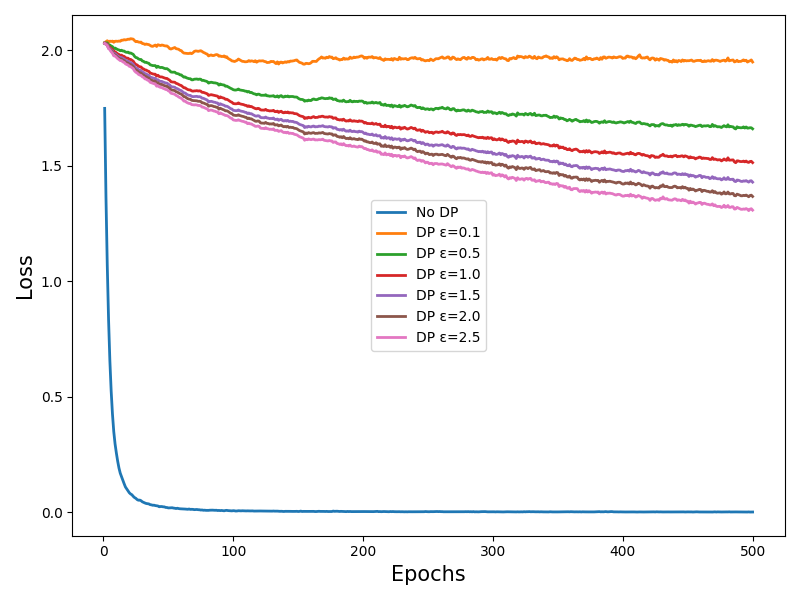}
\\[-0.5em]
% Row 2: MultiModalGCN
\rotatebox{90}{\hspace{1em} \textbf{MultiModalGCN}} &
\includegraphics[width=0.45\linewidth]{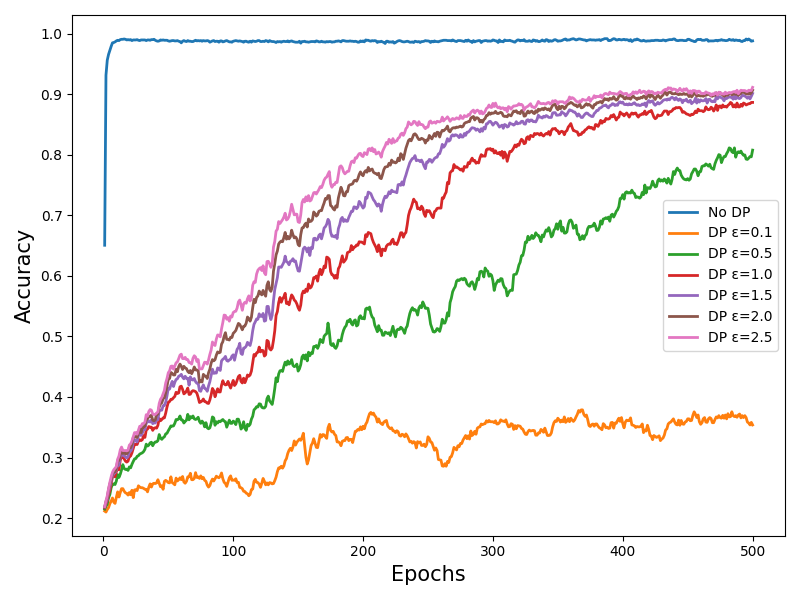} &
\includegraphics[width=0.45\linewidth]{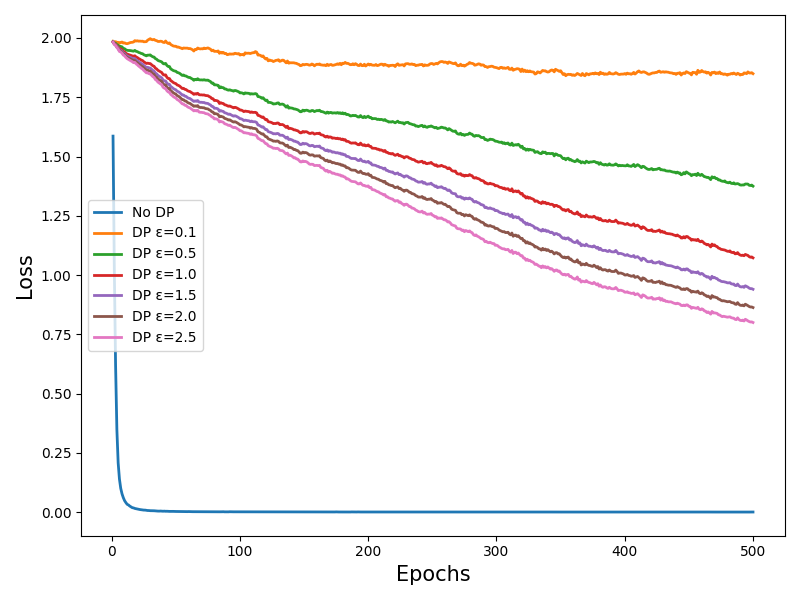}
\end{tabular}
\vspace{0.5em}
\caption{Impact of privacy level ($\varepsilon$) on model convergence in Centralized Learning.}
\label{fig:centralized_accuracy_loss}
\end{figure}
% ========== Federated Learning ==========
\begin{figure}[hbt]
\centering
\setlength{\tabcolsep}{4pt}
\renewcommand{\arraystretch}{1.2}
\begin{tabular}{c c c}
% Header row
& \small Accuracy & \small Loss \\
\midrule
% Row 1: MultiModalFFN
\rotatebox{90}{\hspace{1em} \textbf{MultiModalFFN}} &
\includegraphics[width=0.45\linewidth]{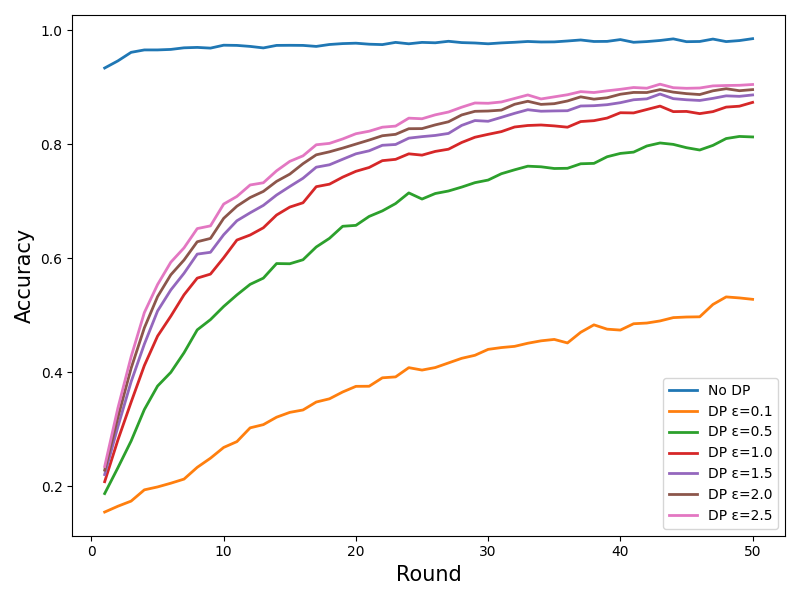} &
\includegraphics[width=0.45\linewidth]{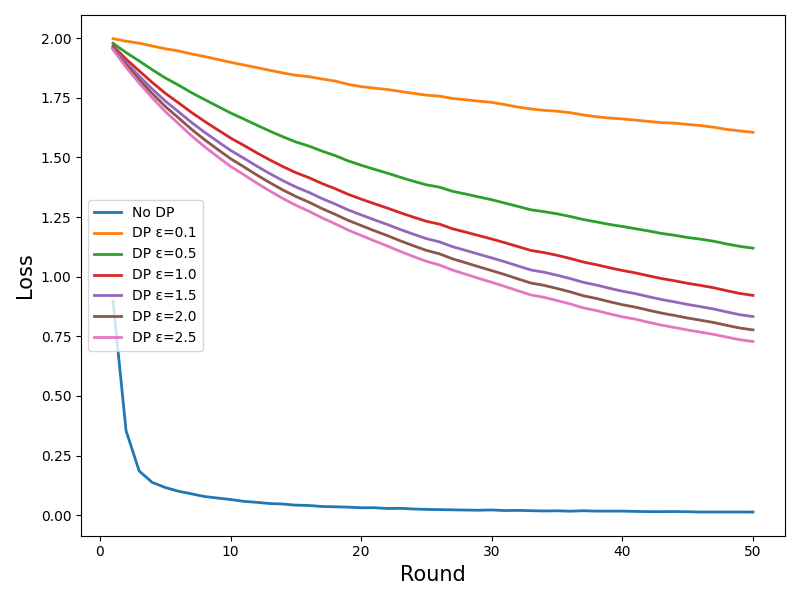}
\\[-0.5em]
% Row 2: MultiModalGCN
\rotatebox{90}{\hspace{1em} \textbf{MultiModalGCN}} &
\includegraphics[width=0.45\linewidth]{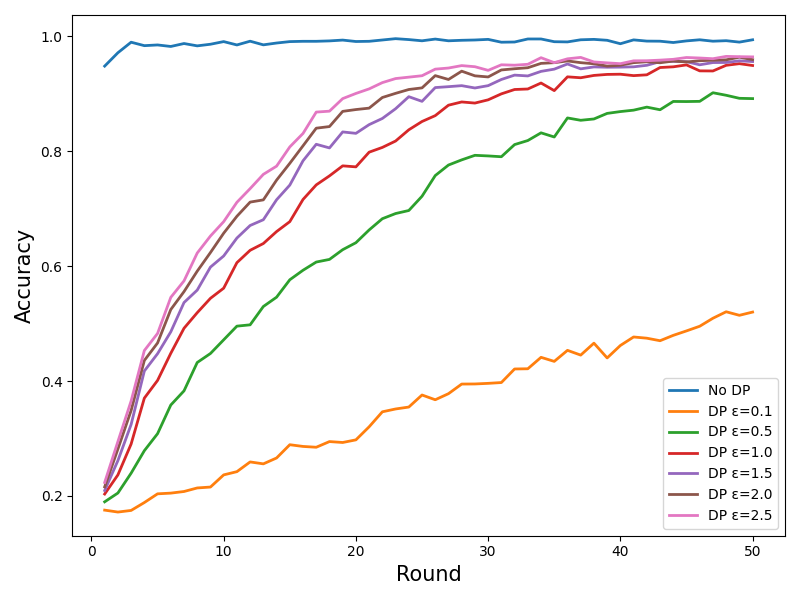} &
\includegraphics[width=0.45\linewidth]{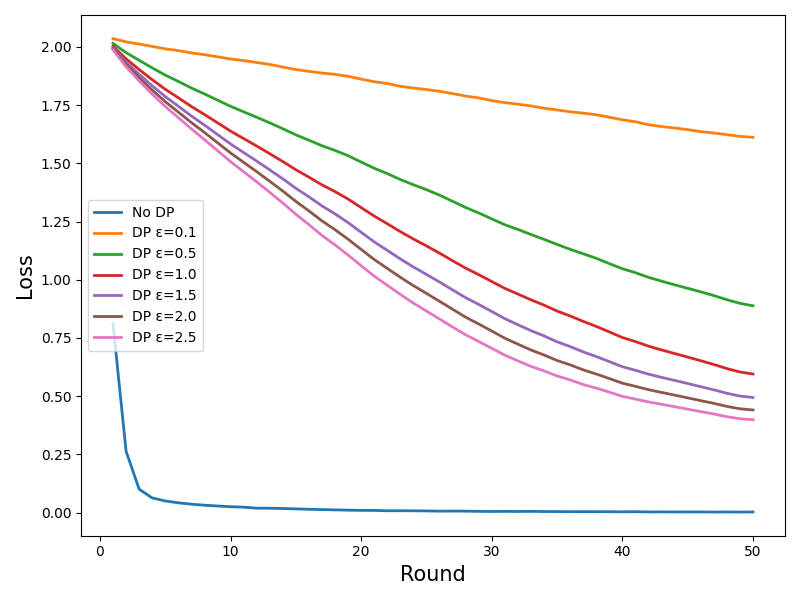}
\end{tabular}
\vspace{0.5em}
\caption{Impact of privacy level ($\varepsilon$) on model convergence in Federated Learning.}
\label{fig:federated_accuracy_loss}
\end{figure}
Figure~(\ref{fig:centralized_accuracy_loss}) and (\ref{fig:federated_accuracy_loss}) clearly illustrate a strong privacy-utility trade-off: tighter privacy (smaller $\varepsilon$) significantly slows convergence and reduces accuracy, while larger $\varepsilon$ narrows the gap with non-DP baselines. In the federated learning, models without DP achieves $>95\%$ accuracy within a few rounds, whereas DP enforced models show $\varepsilon$ dependent behavior $\varepsilon=2.5$ reaches ~$90\%$, but $\varepsilon=0.1$ struggles to exceed $50\%$. The centralized setting magnifies these effects models improve gradually and more oscillatory convergence, while federated models stabilize early. Across both settings, GCN proves more resilient than FFN. The key insights are model architecture and training paradigm strongly influence the privacy performance balance.
\subsubsection{Comparative Evaluation: Graph-based vs FeedForward Architectures with DP }
\begin{figure}[hbt]
\centering
\setlength{\tabcolsep}{3pt} % reduce column spacing a bit
\renewcommand{\arraystretch}{1} % normal row spacing
\begin{tabular}{c c c}
% Header Row
& \small F1 Score & \small Loss \\
\midrule
% Row 1: Centralized
\rotatebox{90}{\hspace{0.8em} \textbf{Centralized}} &
\includegraphics[width=0.45\linewidth]{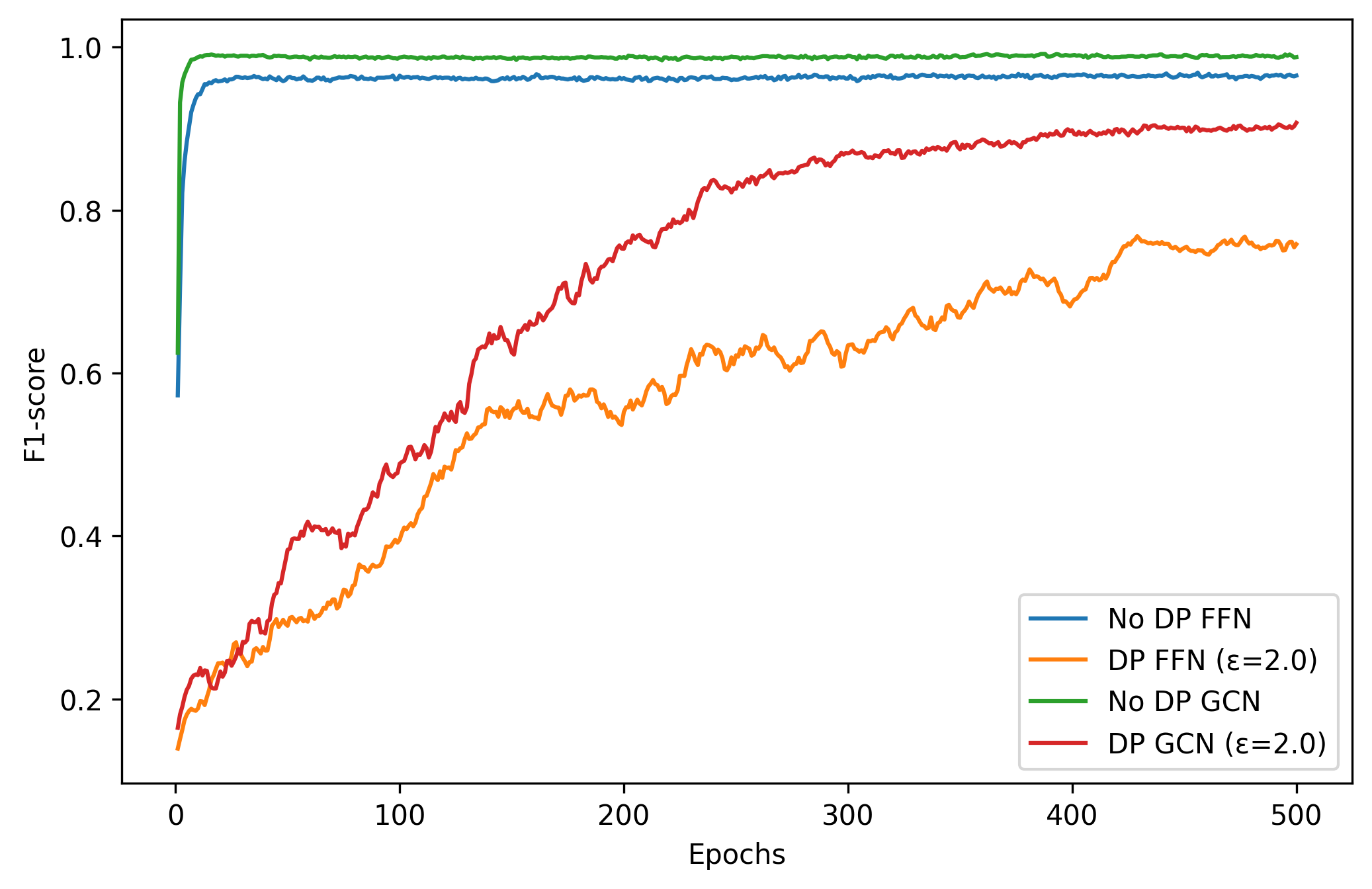} &
\includegraphics[width=0.45\linewidth]{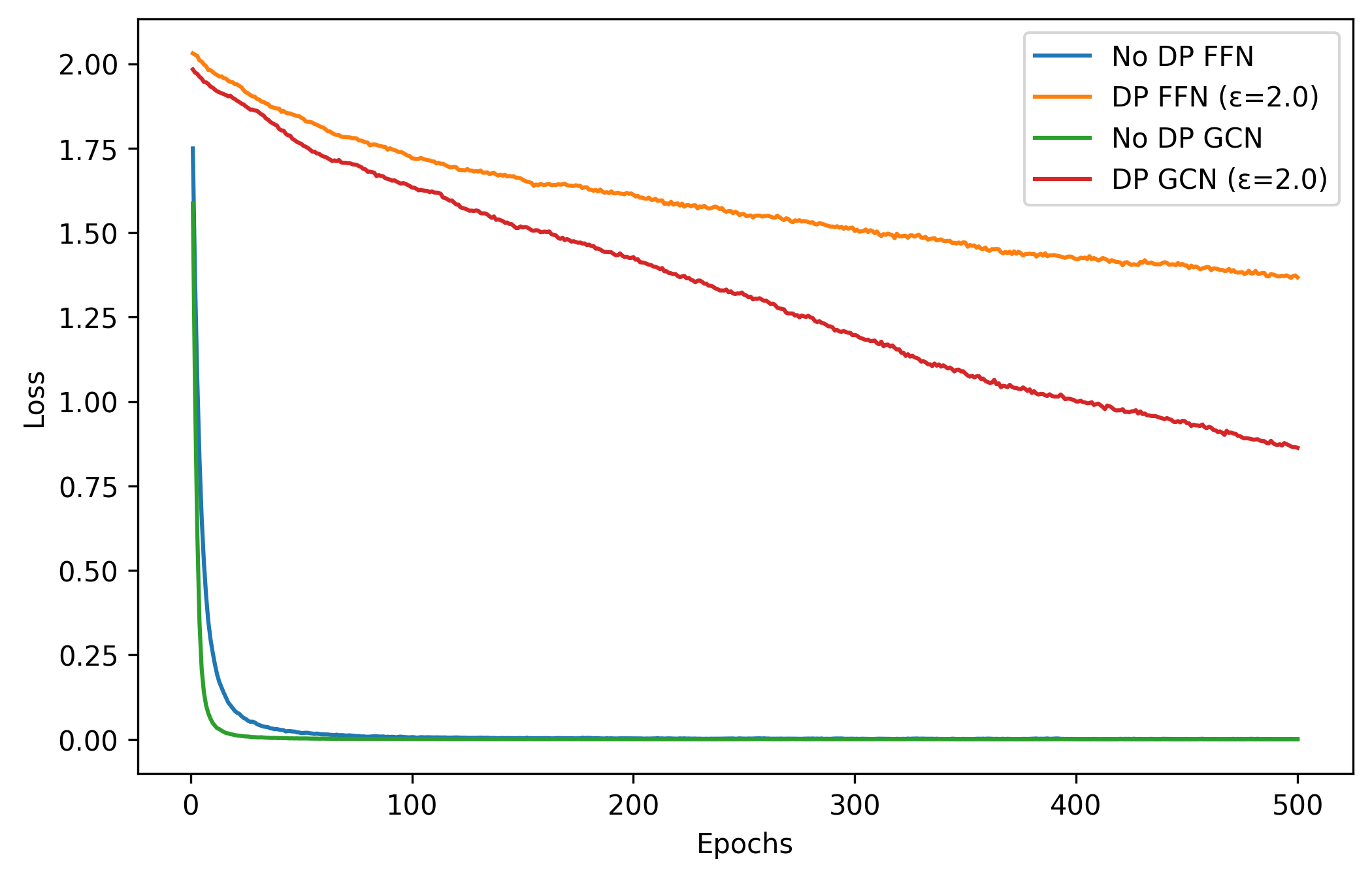}
\\[-0.5em]
% Row 2: Federated
\rotatebox{90}{\hspace{0.8em} \textbf{Federated}} &
\includegraphics[width=0.45\linewidth]{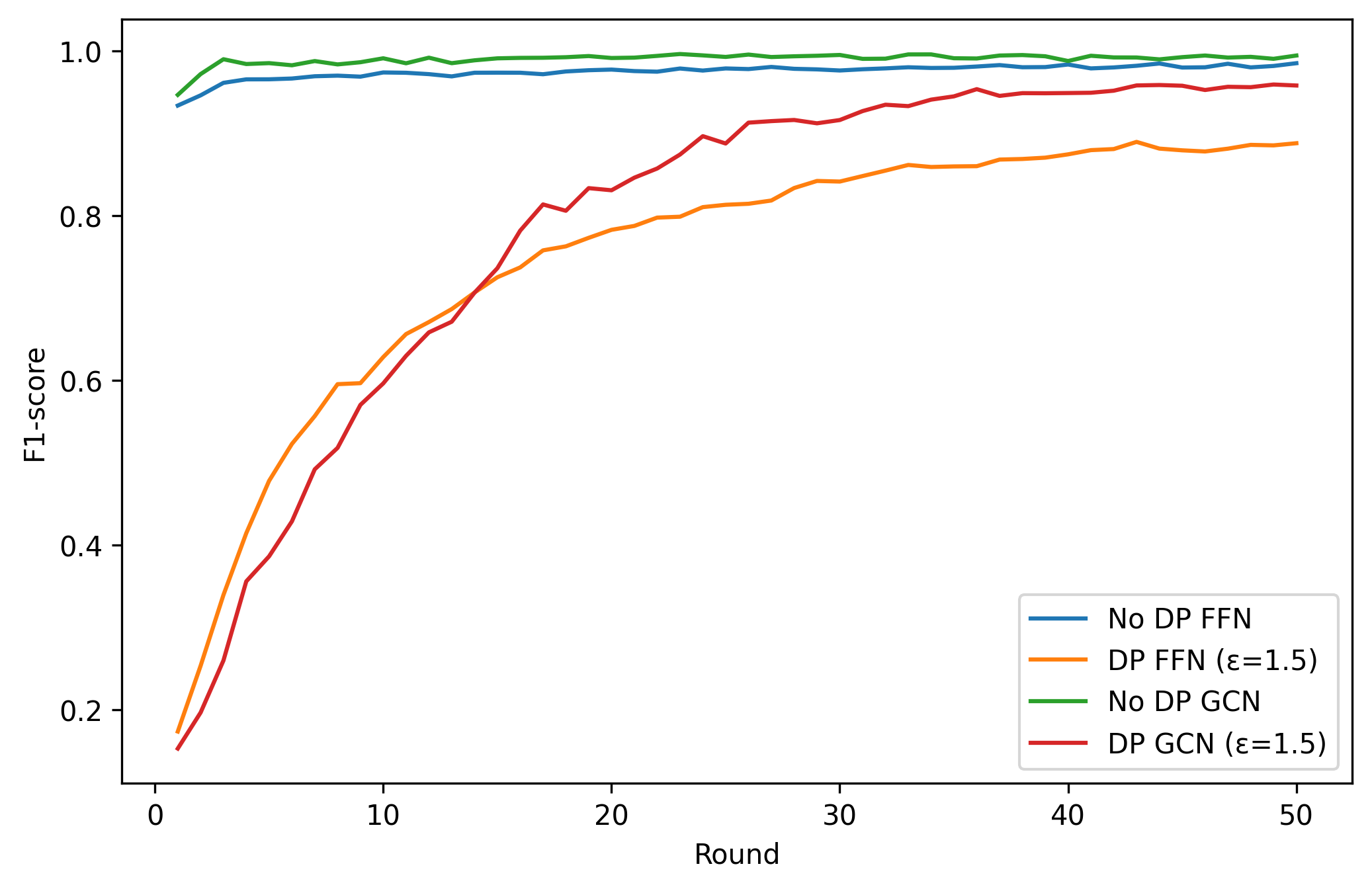} &
\includegraphics[width=0.45\linewidth]{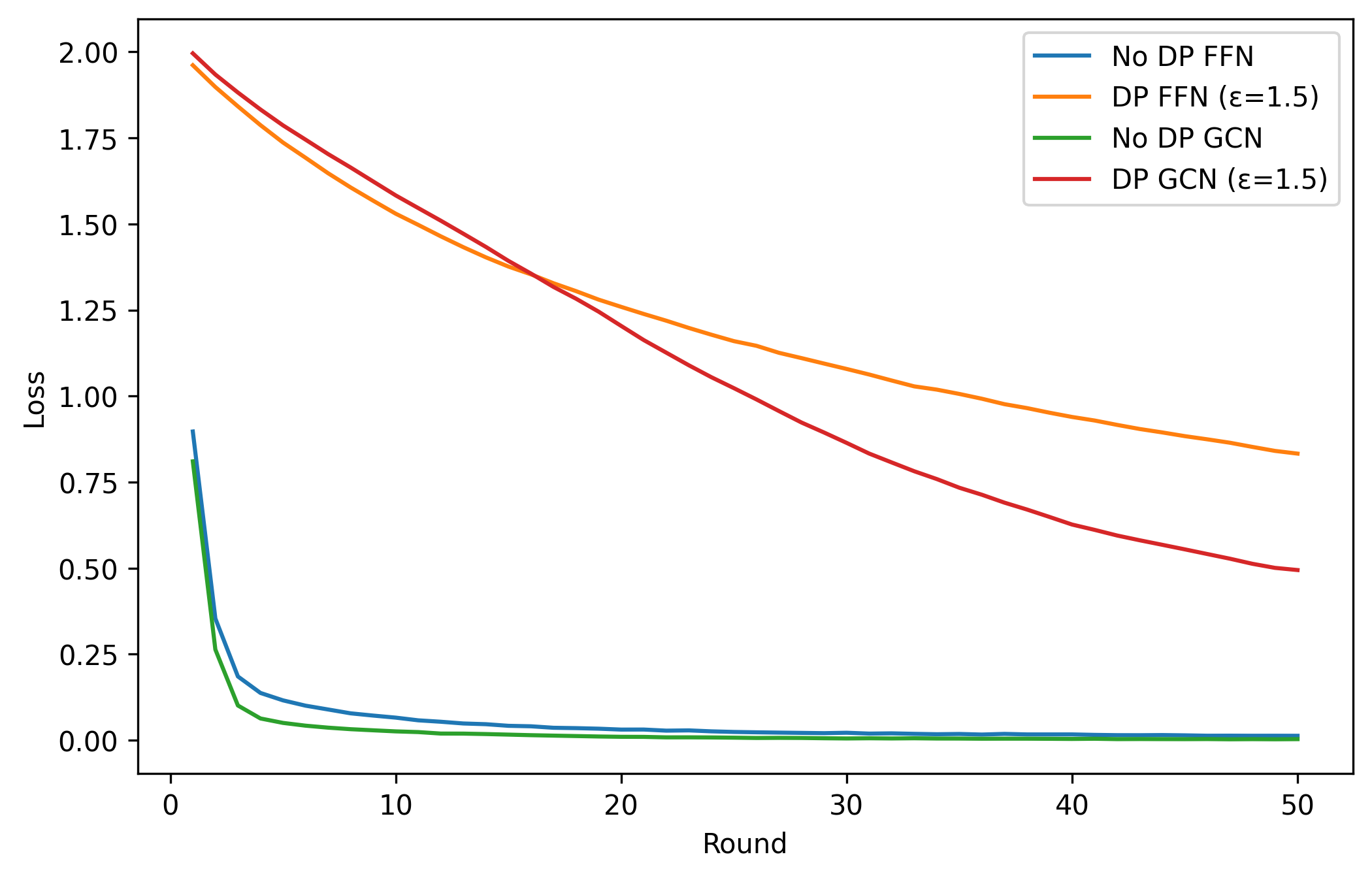}
\end{tabular}
\vspace{0.5em}
\caption{Performance metrics under Weak DP ($\varepsilon=2.0, \sigma=0.7852$) for centralized and federated models.}
\label{fig:weak_dp_compare}
\end{figure}
% ==================== Strong DP ====================
\begin{figure}[hbt]
\centering
\setlength{\tabcolsep}{3pt}
\renewcommand{\arraystretch}{1}
\begin{tabular}{c c c}
% Header Row
& \small F1 Score & \small Loss \\
\midrule
% Row 1: Centralized
\rotatebox{90}{\hspace{0.8em} \textbf{Centralized}} &
\includegraphics[width=0.45\linewidth]{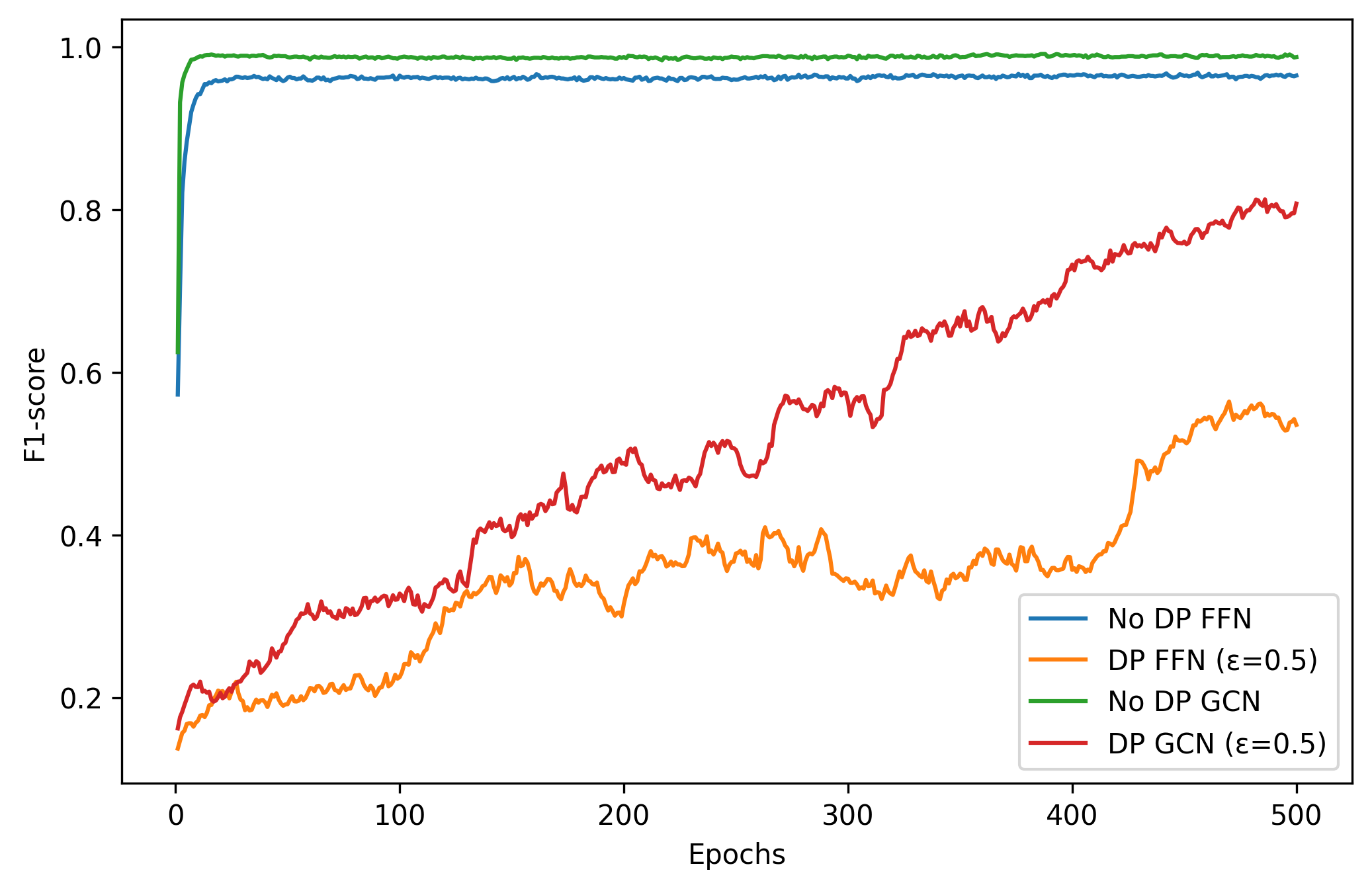} &
\includegraphics[width=0.45\linewidth]{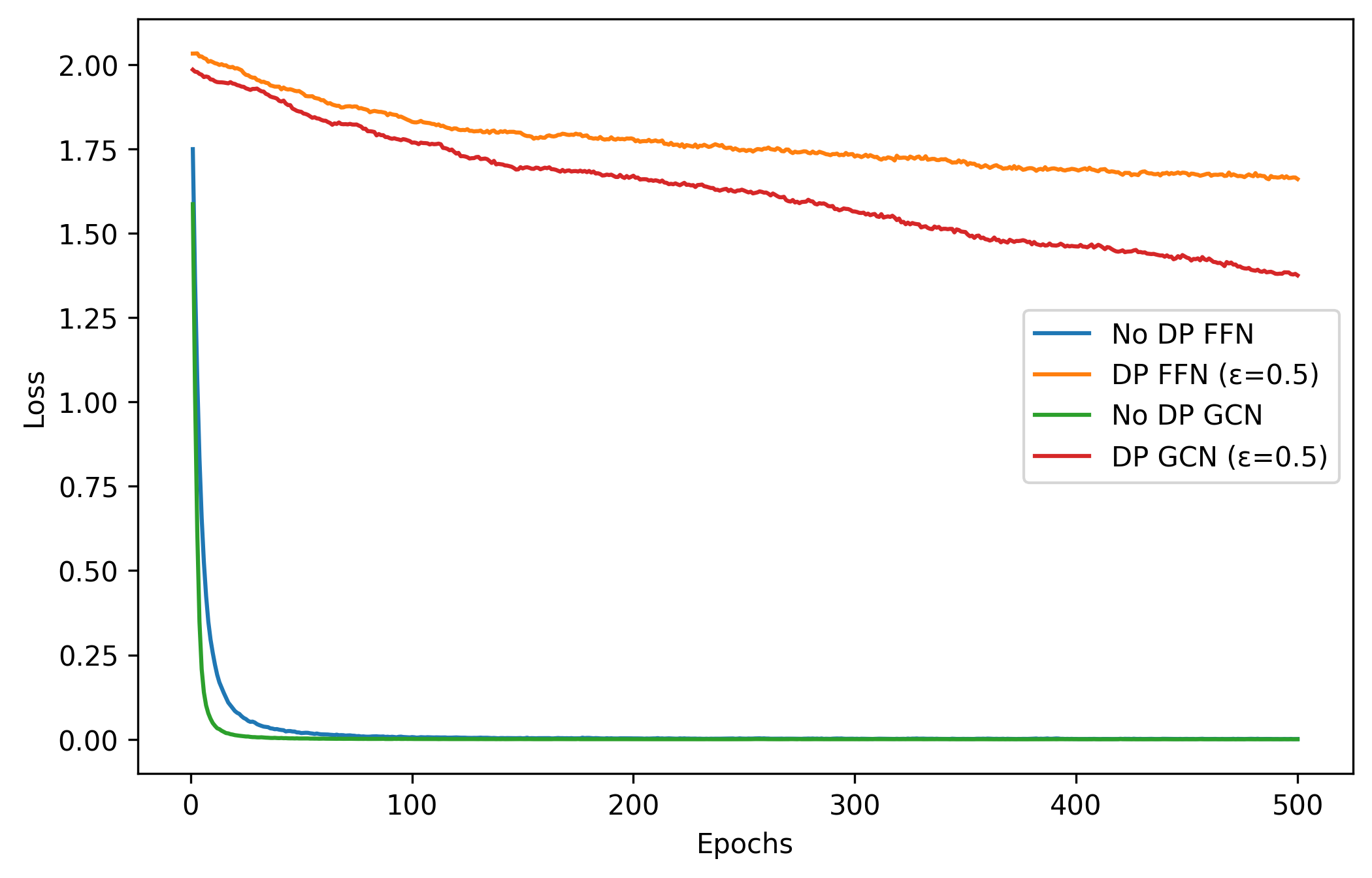}
\\[-0.5em]
% Row 2: Federated
\rotatebox{90}{\hspace{0.8em} \textbf{Federated}} &
\includegraphics[width=0.45\linewidth]{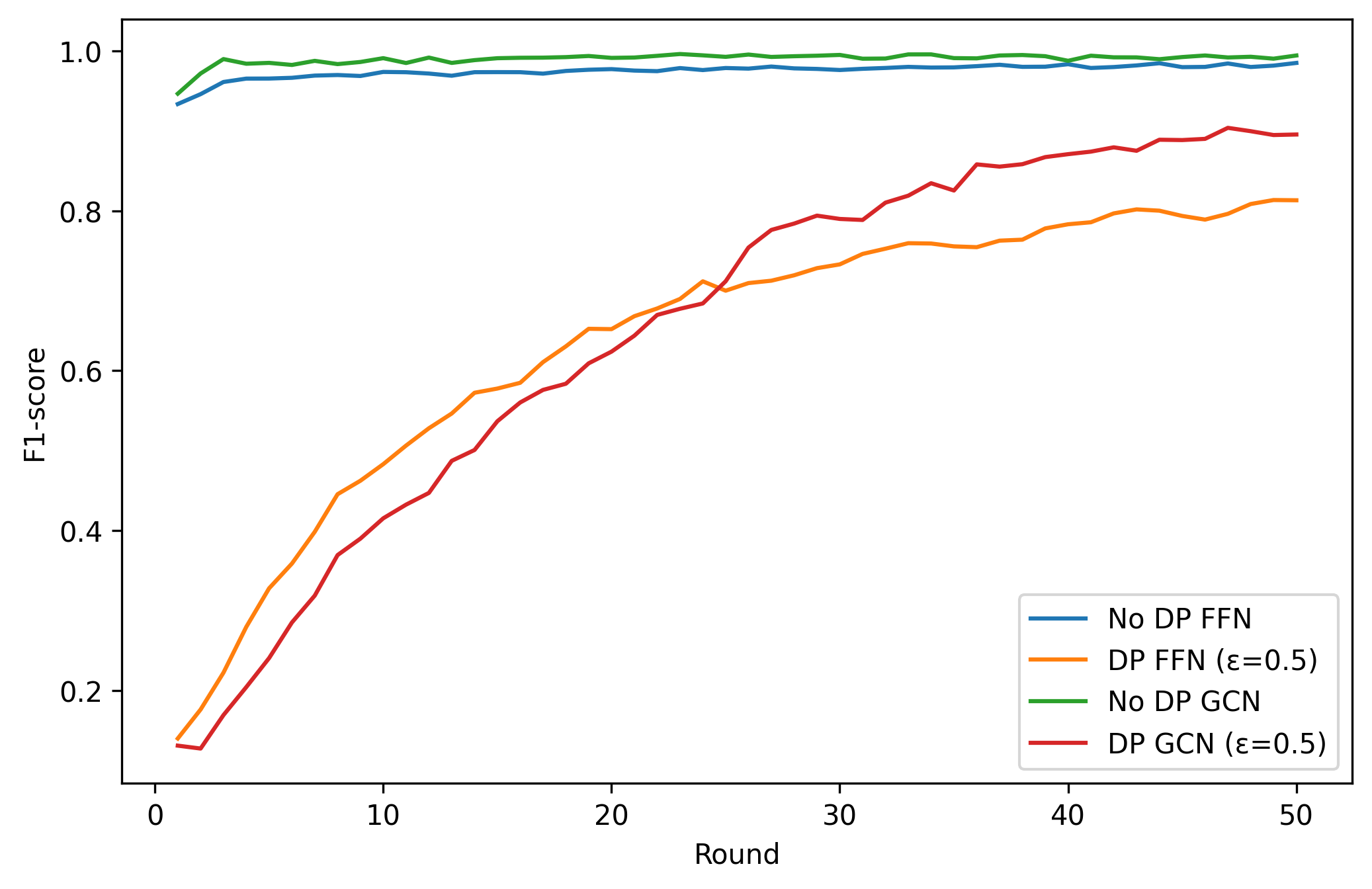} &
\includegraphics[width=0.45\linewidth]{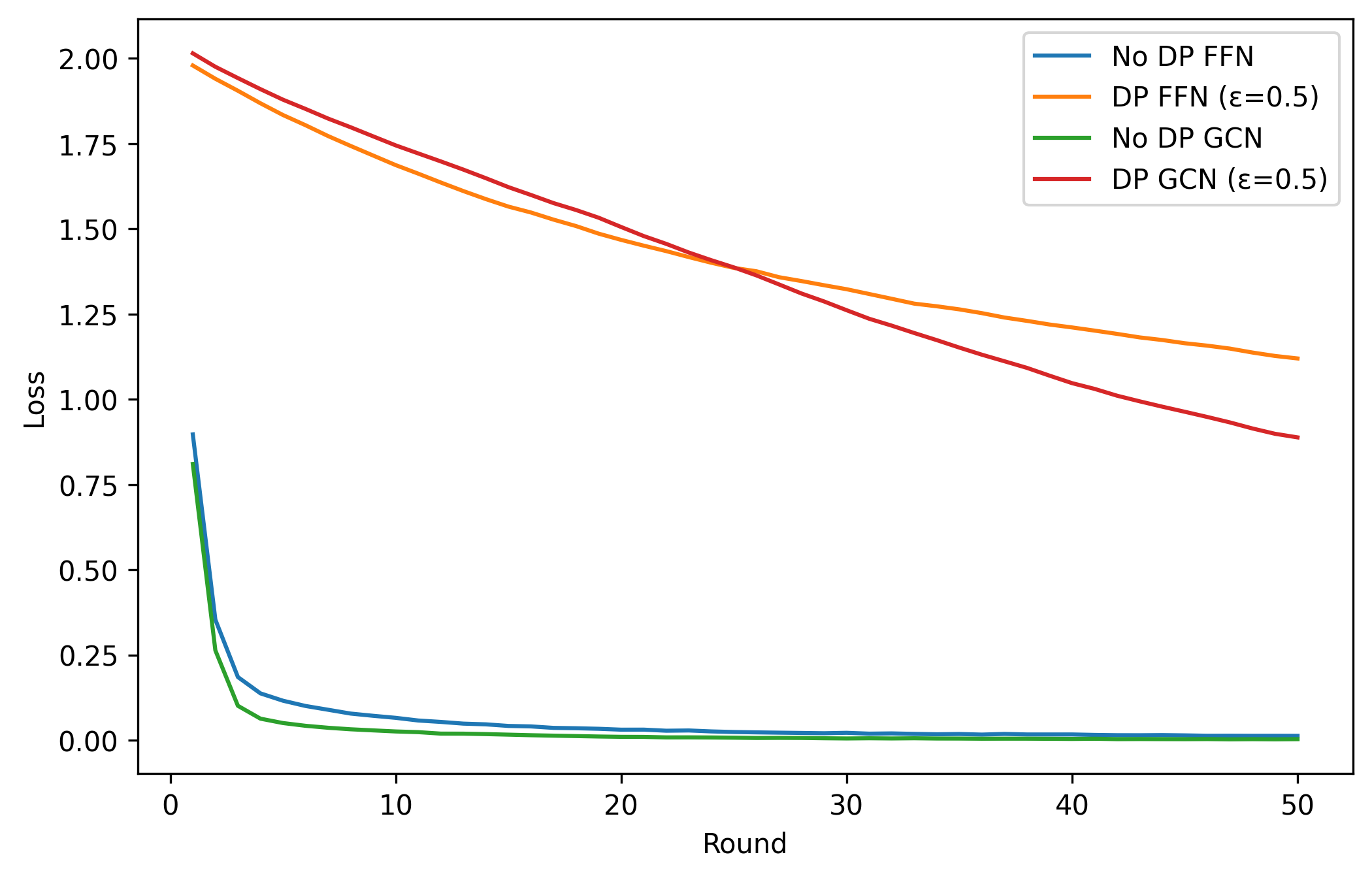}
\end{tabular}
\vspace{0.5em}
\caption{Performance metrics under Strong DP ($\varepsilon=0.5, \sigma=1.4216$) for centralized and federated models.}
\label{fig:strong_dp_compare}
\end{figure}
Figures~(\ref{fig:weak_dp_compare}) and (\ref{fig:strong_dp_compare}) provide a comparative analysis of graph-based and feedforward models under strong and weak DP cases. In centralized learning, GCN consistently outperforms than FFN. Under federated paradigm, FFN shows a slight early advantage, surpassing GCN only up to ~$15$ rounds for weak DP and ~$26$ rounds for strong DP, after which GCN dominates, highlighting its superiority to privacy-induced noise and sustained performance.
\subsubsection{Utility Loss}
\begin{table}[hbt]
\centering
\caption{Utility loss distributions across privacy budgets for centralized vs.\ federated training. $500$ epochs for centralized and $50$ rounds for federated}
\label{tab:utility_comparison}
\begin{tabular}{c c}
\toprule
\textbf{Centralized} & \textbf{Federated} \\
\midrule
\includegraphics[width=0.45\linewidth]{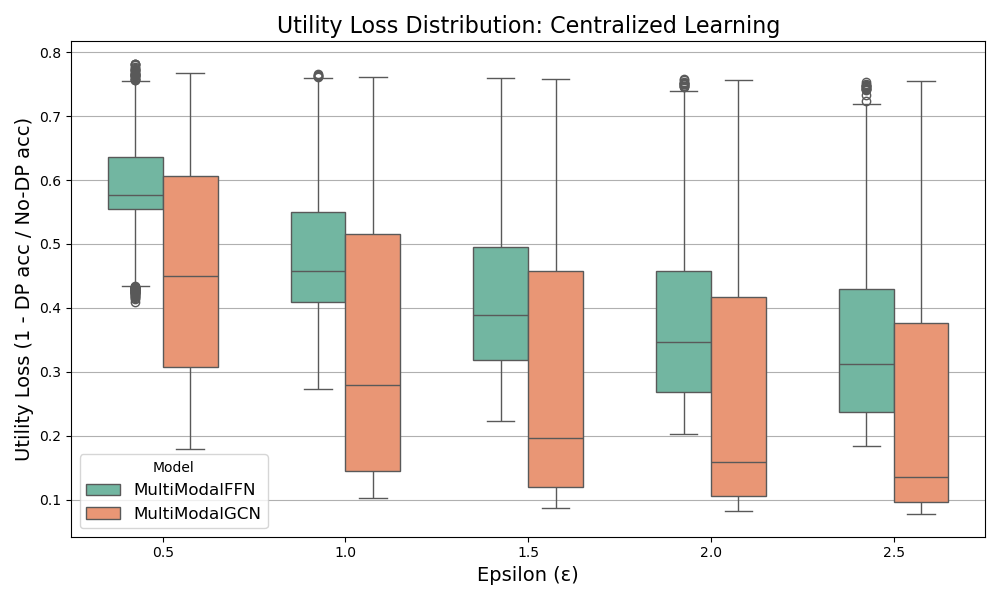} &
\includegraphics[width=0.45\linewidth]{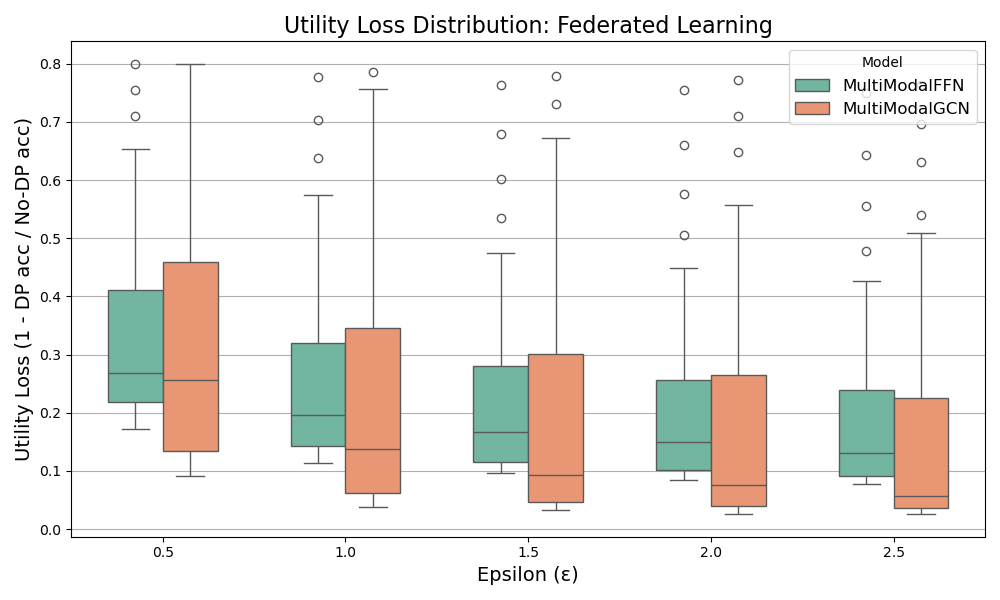} \\
% \parbox{0.45\linewidth}{\centering Learning stability under varying privacy budgets (No-DP, $\varepsilon{=}0.5,1.0,1.5,2.0,2.5$), showing median and variability (min–max) across $500$ epochs.} &
% \parbox{0.45\linewidth}{\centering Learning dynamics with differential privacy budgets (No-DP, $\varepsilon{=}0.5,1.0,1.5,2.0,2.5$), highlighting performance variations via median and min–max ranges across $50$ rounds.} \\
\bottomrule
\end{tabular}
\end{table}
Table~III presents boxplots of the Utility Loss  for Federated and Centralized learning. It is observed that federated learning outperforms centralized learning under strict privacy budgets, demonstrating resilience to DP noise, but centralized learning delivers superior stability and consistency as privacy constraints relax. This contrast highlights the trade-off: federated learning is stronger for privacy while centralized learning is stronger for robustness.
\subsubsection{Ablation Study on Fraction of Client's Participation at Each Round}
\begin{figure}[hbt]
    \centering
    \begin{subfigure}[b]{0.48\linewidth}
        \includegraphics[width=\linewidth]{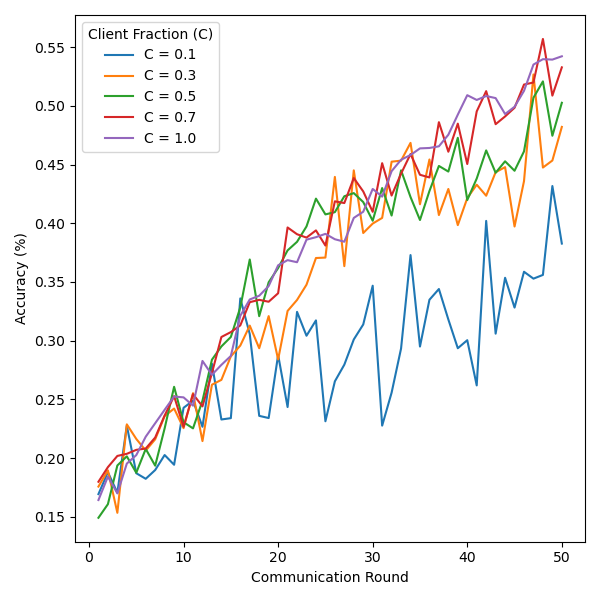}
        \caption*{Strong Privacy Budget}
    \end{subfigure}
    \hfill
    \begin{subfigure}[b]{0.48\linewidth}
        \includegraphics[width=\linewidth]{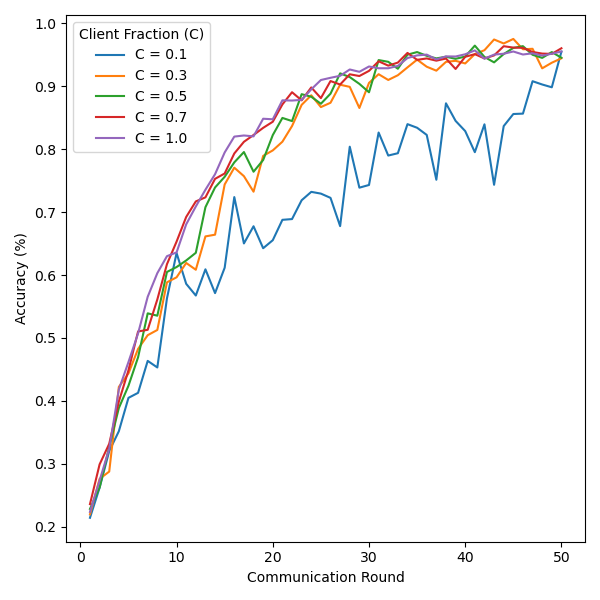}
        \caption*{Weak Privacy Budget}
    \end{subfigure}
    \caption{Effect of client participation on accuracy under DP.}
    \label{fig:sidebyside}
\end{figure}
Figure~(\ref{fig:sidebyside}) under strong DP budgets, increasing the fraction of participating clients significantly boosts accuracy by averaging out the amplified noise, whereas under weak DP, client's participation has a smaller marginal effect since the noise impact is already reduced. This highlights that high client participation is crucial for utility preservation when privacy guarantees are strict, but less critical when privacy is relaxed.
\subsubsection{Ablation study of Modality Combinations Across Privacy Levels and Learning setting}
\begin{table}[hbt!]
\centering
\caption{Ablation Study of Three Modality Combinations}
\begin{tabular}{p{1cm} c c c c c c}
\toprule
\multirow{2}{*}{Modality} & \multirow{2}{*}{DP Setting} & \multicolumn{2}{c}{Centralized} & \multicolumn{2}{c}{Federated} \\
\cmidrule(lr){3-4} \cmidrule(lr){5-6}
 & & Accuracy & F1 & Accuracy & F1 \\
\midrule
\multirow{3}{*}{act, acw, dc} 
 & $\varepsilon=\infty$ &  0.9867  & 0.9869 & \textbf{0.9925} & \textbf{0.9928} \\
 & $\varepsilon=0.5$  & 0.8143 & 0.8108 & \textbf{0.9363} & \textbf{0.9385} \\
 & $\varepsilon=2.0$  & 0.9069 & 0.9080 & \textbf{0.9576} & \textbf{0.9595} \\
\midrule
\multirow{3}{*}{act, acw, pm} 
 & $\varepsilon=\infty$    & \textbf{0.9905} & 0.9907 & 0.9905 & \textbf{0.9908} \\
 & $\varepsilon=0.5$  & 0.8146 & 0.8114 & \textbf{0.9314} & \textbf{0.9335} \\
 & $\varepsilon=2.0$  & 0.9120 & 0.9134 & \textbf{0.9575} & \textbf{0.9595} \\
\midrule
\multirow{3}{*}{act, dc, pm} 
 & $\varepsilon=\infty$    & 0.9814 & 0.9816 & \textbf{0.9917} & \textbf{0.9917} \\
 & $\varepsilon=0.5$  & 0.7377 & 0.7327 & \textbf{0.8970} & \textbf{0.8981} \\
 & $\varepsilon=2.0$  & 0.8894 & 0.8989 & \textbf{0.9450} & \textbf{0.9466} \\
\midrule
\multirow{3}{*}{acw, dc, pm} 
 & $\varepsilon=\infty$    & \textbf{0.9814} & \textbf{0.9816} & 0.9192 & 0.9194 \\
 & $\varepsilon=0.5$  & \textbf{0.7440} & \textbf{0.7397} & 0.6740 & 0.6743 \\
 & $\varepsilon=2.0$  & \textbf{0.8921} & \textbf{0.8942} & 0.7569 & 0.7585 \\
\bottomrule
\end{tabular}
\label{tab:modality_performance_1}
\end{table}
\begin{table}[hbt!]
\centering
\caption{Ablation Study of Two Modality Combinations}
\begin{tabular}{p{1cm} c c c c c c}
\toprule
\multirow{2}{*}{Modality} & \multirow{2}{*}{DP Setting} & \multicolumn{2}{c}{Centralized} & \multicolumn{2}{c}{Federated} \\
\cmidrule(lr){3-4} \cmidrule(lr){5-6}
 & & Accuracy & F1 & Accuracy & F1 \\
\midrule
\multirow{3}{*}{act, acw} 
 & $\varepsilon=\infty$    & 0.9874 & 0.9875 & \textbf{0.9909} & \textbf{0.9912} \\
 & $\varepsilon=0.5$  & 0.8837 & 0.8824 & \textbf{0.9282} & \textbf{0.9293} \\
 & $\varepsilon=2.0$  & 0.9445 & 0.9454 & \textbf{0.9635} & \textbf{0.9644} \\
\midrule
\multirow{3}{*}{act, dc} 
 & $\varepsilon=\infty$    & 0.9799 & 0.9800 & \textbf{0.9846} & \textbf{0.9848} \\
 & $\varepsilon=0.5$  & 0.8090 & 0.8067 & \textbf{0.9280} & \textbf{0.9294} \\
 & $\varepsilon=2.0$  & 0.9016 & 0.9025 & \textbf{0.9499} & \textbf{0.9505} \\
\midrule
\multirow{3}{*}{act, pm} 
 & $\varepsilon=\infty$    & 0.9546  & 0.9521 & \textbf{0.9794} & \textbf{0.9797} \\
 & $\varepsilon=0.5$  & 0.8445 & 0.8498 & \textbf{0.9207} & \textbf{0.9231} \\
 & $\varepsilon=2.0$  & 0.8999 & 0.8998 & \textbf{0.9453} & \textbf{0.9466} \\
\midrule
\multirow{3}{*}{acw, dc} 
 & $\varepsilon=\infty$   & \textbf{0.9190} & \textbf{0.9202} & 0.8898 & 0.8905 \\
 & $\varepsilon=0.5$  & 0.6583 & 0.6186 & \textbf{0.6680} & \textbf{0.6723} \\
 & $\varepsilon=2.0$  & 0.6927 & 0.6684 & \textbf{0.7303} & \textbf{0.7347} \\
\midrule
\multirow{3}{*}{acw, pm} 
 & $\varepsilon=\infty$    & \textbf{0.9062} & \textbf{0.9055} & 0.8830 & 0.8827 \\
 & $\varepsilon=0.5$  & \textbf{0.6993} & \textbf{0.6998} & 0.6523 & 0.6526 \\
 & $\varepsilon=2.0$  & \textbf{0.7898} & \textbf{0.7998} & 0.7024 & 0.7058 \\
\midrule
\multirow{3}{*}{dc, pm} 
 & $\varepsilon=\infty$    & 0.9856 & 0.9875 & \textbf{0.9940} & \textbf{0.9938} \\
 & $\varepsilon=0.5$  & 0.8355 & 0.8299 & \textbf{0.8417} & \textbf{0.8424} \\
 & $\varepsilon=2.0$  & 0.8963 & 0.8974 & \textbf{0.9108} & \textbf{0.9124} \\
\midrule
\bottomrule
\end{tabular}
\label{tab:modality_performance_2}
\end{table}
\begin{table}[hbt]
\centering
\caption{Ablation Study of Uni-modality}
\begin{tabular}{p{1cm} c c c c c c}
\toprule
\multirow{2}{*}{Modality} & \multirow{2}{*}{DP Setting} & \multicolumn{2}{c}{Centralized} & \multicolumn{2}{c}{Federated} \\
\cmidrule(lr){3-4} \cmidrule(lr){5-6}
 & & Accuracy & F1 & Accuracy & F1 \\
\midrule
\multirow{3}{*}{act} 
 & $\varepsilon=\infty$    & \textbf{0.9806} & \textbf{0.9807} & 0.9787 & 0.9790\\
 & $\varepsilon=0.5$  & 0.8214 & 0.8209 & \textbf{0.9134} & \textbf{0.9152}\\
 & $\varepsilon=2.0$  & 0.9068 & 0.9083 & \textbf{0.9457}  & \textbf{0.9469} \\
\midrule
\multirow{3}{*}{acw} 
 & $\varepsilon=\infty$    & \textbf{0.9136} & \textbf{0.9152} & 0.8457 & 0.8489\\
 & $\varepsilon=0.5$  & 0.6099 & 0.5848 & \textbf{0.6434} & \textbf{0.6419}\\
 & $\varepsilon=2.0$  & 0.6947 & 0.6962 & \textbf{0.7211} & \textbf{0.7227} \\
\midrule
\multirow{3}{*}{dc} 
 & $\varepsilon=\infty$    & \textbf{0.9968} & \textbf{0.9967} & 0.9947 & 0.9946 \\
 & $\varepsilon=0.5$  & 0.8672 & 0.8696 & \textbf{0.8695} & \textbf{0.8704} \\
 & $\varepsilon=2.0$  & 0.9000 & 0.9018 & \textbf{0.9231} & \textbf{0.9243}\\
\midrule
\multirow{3}{*}{pm} 
 & $\varepsilon=\infty$    & \textbf{0.7088} & \textbf{0.6998} & 0.6902 & 0.6728\\
 & $\varepsilon=0.5$  & \textbf{0.3856} & \textbf{0.3896} & 0.3748 &0.3219 \\
 & $\varepsilon=2.0$  & \textbf{0.4299} & \textbf{0.4395} & 0.4146 & 0.3766\\
\bottomrule
\end{tabular}
\label{tab:modality_performance_3}
\end{table}
From Table ~(\ref{tab:modality_performance_1}), (\ref{tab:modality_performance_2}), and (\ref{tab:modality_performance_3}) multi-modal combinations significantly boost performance, especially in federated learning under strict privacy constraints. In contrast, single modality model, notably show lower performance, especially pm modality case. Integrating multiple modalities boosts clarity and enriches better understanding about the model.
\subsubsection{The Cost of Privacy: Error Trends in Exercise Classification}
% \begin{figure}[h]
%     \centering
%     % First row
%     \begin{minipage}{0.5\linewidth}
%         \centering
%         \includegraphics[width=\linewidth]{modalityall4_gcn_plots/confusion_matrix_eps0.1.png}
%         \caption*{$\varepsilon = 0.1$}
%     \end{minipage}
%     \hfill
%     \begin{minipage}{0.5\linewidth}
%         \centering
%         \includegraphics[width=\linewidth]{modalityall4_gcn_plots/confusion_matrix_eps0.5.png}
%         \caption*{$\varepsilon = 0.5$}
%     \end{minipage}
%     % \vspace{0.5em}
%     % Second row
%     \begin{minipage}{0.5\linewidth}
%         \centering
%         \includegraphics[width=\linewidth]{modalityall4_gcn_plots/confusion_matrix_eps1.png}
%         \caption*{$\varepsilon = 1.0$}
%     \end{minipage}
%     \hfill
%     \begin{minipage}{0.5\linewidth}
%         \centering
%         \includegraphics[width=\linewidth]{modalityall4_gcn_plots/confusion_matrix_nodp.png}
%         \caption*{$\varepsilon = \infty$ (No DP)}
%     \end{minipage}

%     \caption{Confusion matrices under varying privacy budgets ($\varepsilon$).}
%     \label{fig:conf_matrices_4}
% \end{figure}
\begin{figure}[hbt]
    \centering

    % First row
    \parbox{0.45\linewidth}{
        \centering
        \includegraphics[width=\linewidth]{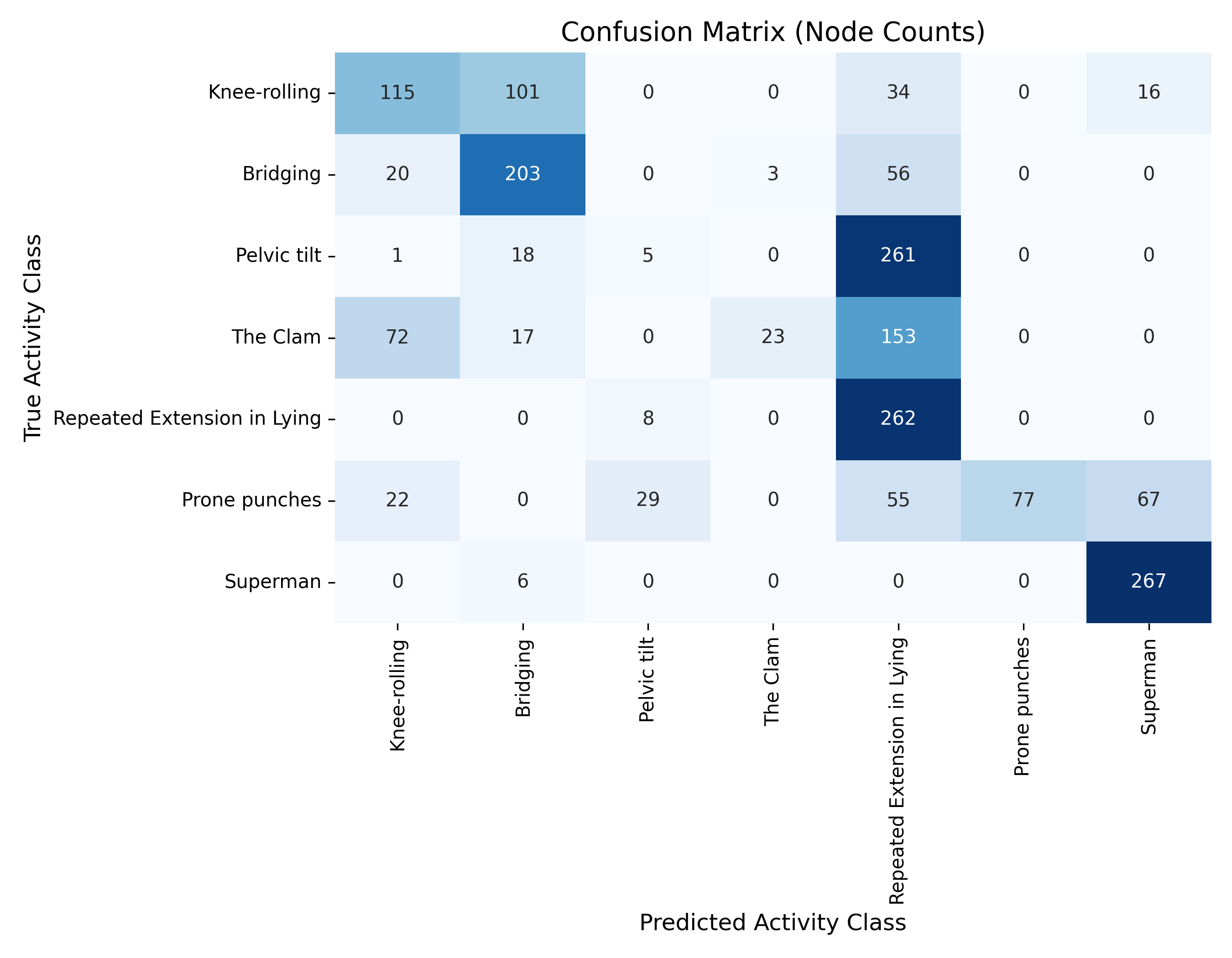}
        \caption*{$\varepsilon = 0.1$}
    }
    \hspace{0.001em}
    \parbox{0.45\linewidth}{
        \centering
        \includegraphics[width=\linewidth]{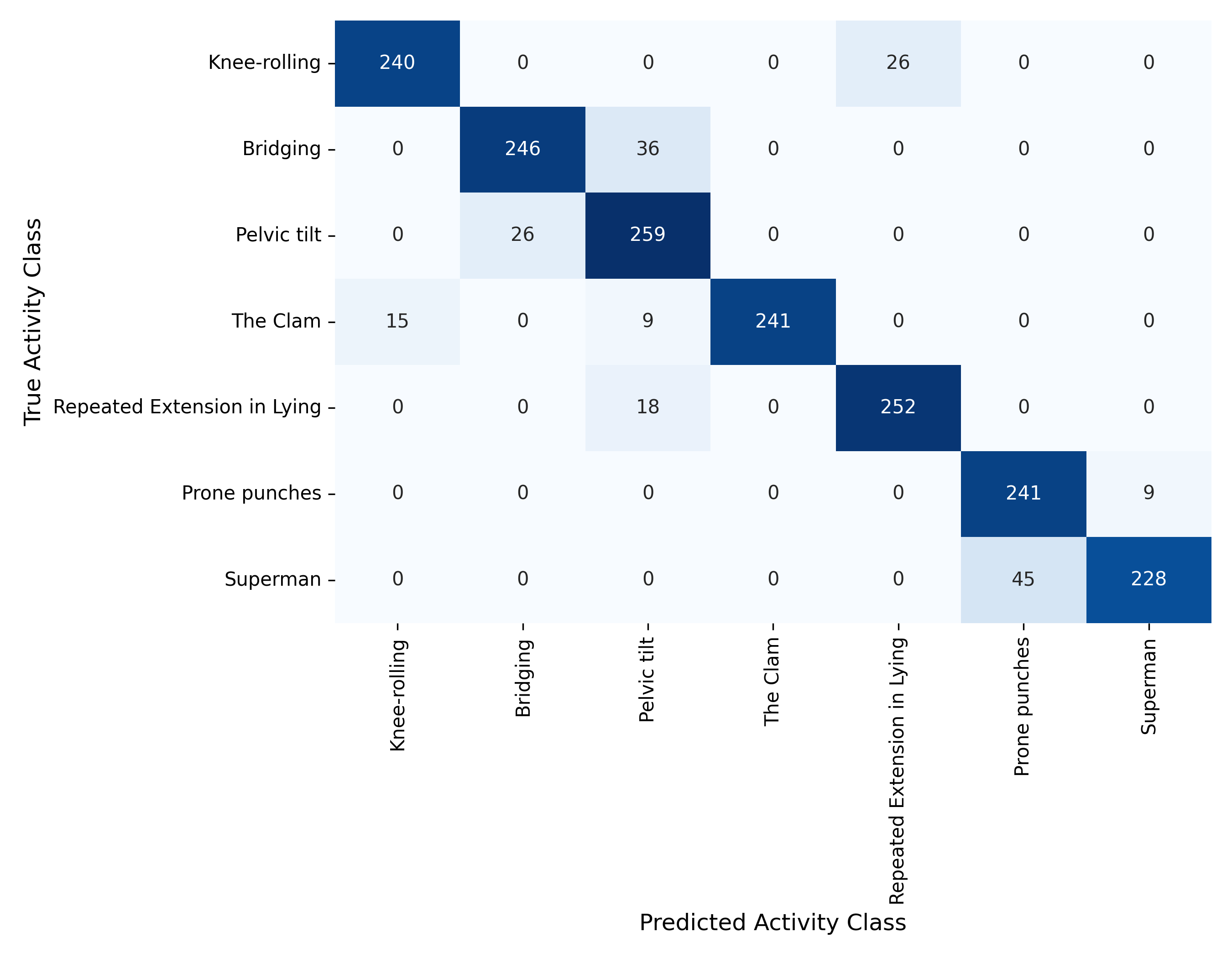}
        \caption*{$\varepsilon = 0.5$}
    }

    % \vspace{1em}

    % Second row
    \parbox{0.45\linewidth}{
        \centering
        \includegraphics[width=\linewidth]{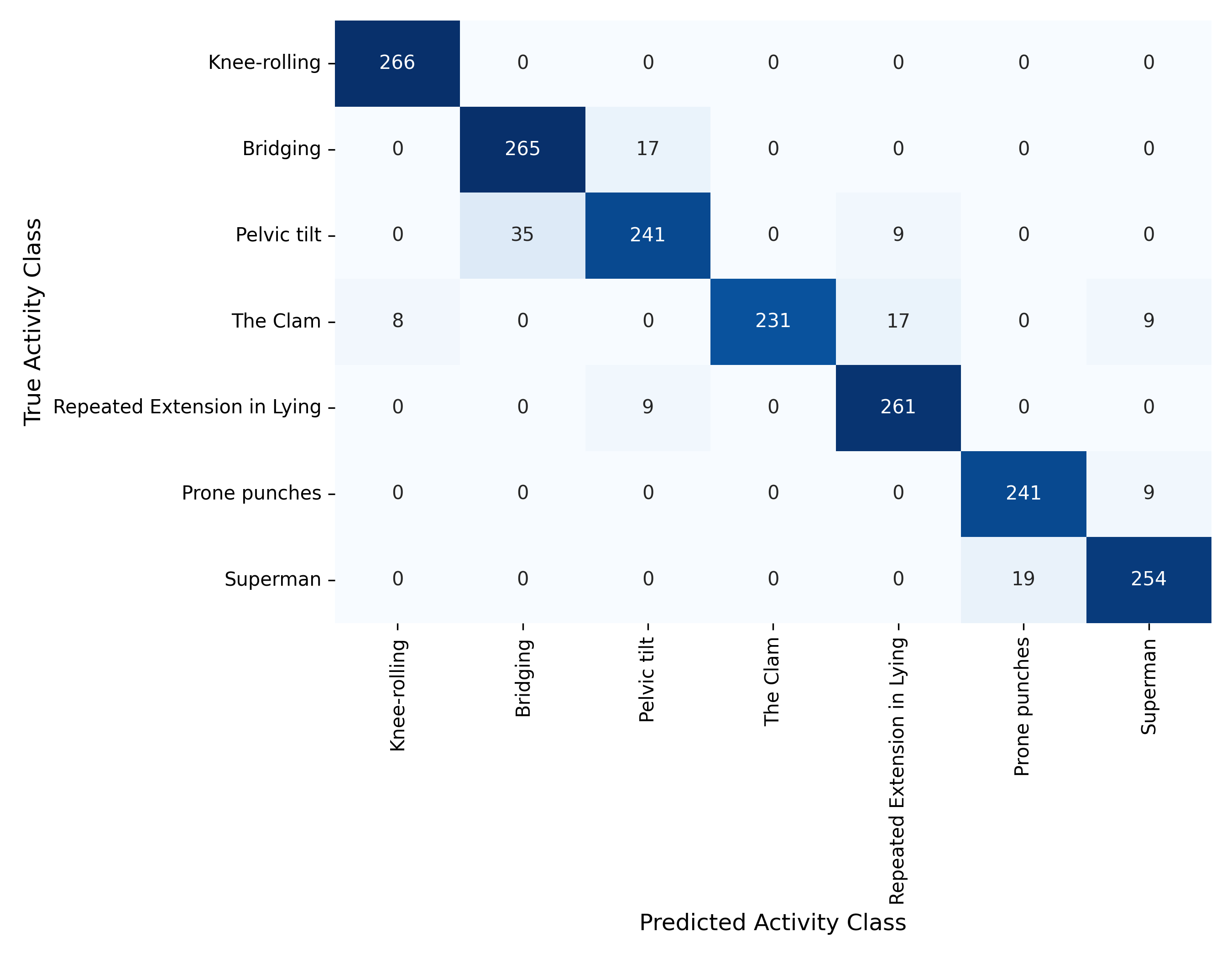}
        \caption*{$\varepsilon = 1.0$}
    }
    \hspace{0.001em}
    \parbox{0.45\linewidth}{
        \centering
        \includegraphics[width=\linewidth]{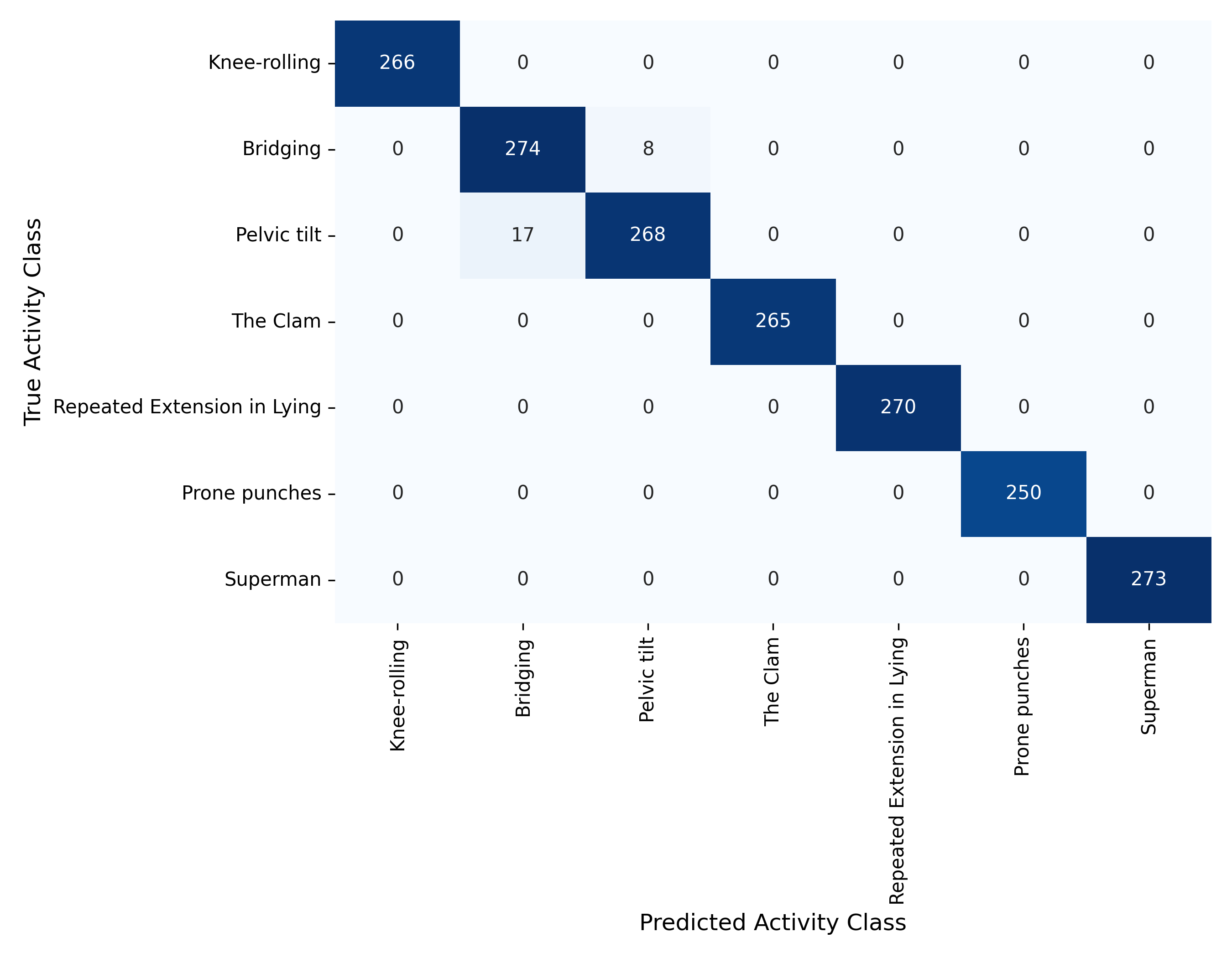}
        \caption*{$\varepsilon = \infty$ (No DP)}
    }

    \caption{Confusion matrices under varying privacy budgets ($\varepsilon$).}
    \label{fig:conf_matrices_4}
\end{figure}

To better understand the trade-off between privacy and utility in federated learning, Figure~(\ref{fig:conf_matrices_4}) shows the confusion matrices under varying privacy budgets, revealing how DP noise reshapes class-level errors.
\begin{itemize}
    \item {$\varepsilon = 0.1$ (Strong DP):} Accuracy collapses as heavy noise erases fine-grained motion cues; subtle exercise (Pelvic tilt, Clam, Prone punches) collapse into dominant classes like Repeated Extension.
    \item {$\varepsilon = 0.5$ (Moderate DP):} Accuracy improves, but residual noise still blurs biomechanically similar classes,e.g., Bridging$\leftrightarrow$ Pelvic tilt.
    \item {$\varepsilon = 1$ (Weak DP):} Errors shrink to a few intrinsic confusions (Superman $\leftrightarrow $ Prone punches); reflects genuine movement similarity rather than privacy noise.
    \item $\varepsilon = \infty$ (No DP): Clear class separation, proving accuracy loss stems from DP noise rather than model limits.
\end{itemize}
% \section{Conclusion}
\newpage
\section{Conclusion}
This article presented GraMFedDHAR, a graph-based multimodal differentially private federated learning framework for privacy-preserving human activity recognition (HAR). Sensor modalities were embedded using DCT (for act, acw) and autoencoder-based method (for dc, pm). This embedded modalities were then modeled as modality-specific graphs and processed through residual GCNs layers, after which their embeddings were fused via an attention-based mechanism. The fused representations were used for GNN-based activity classification at the client level. Integrating differential privacy into the federated training process ensures strong privacy guarantees while mitigating the risk of information leakage. Extensive experiments demonstrated that GCNs are more resilient to privacy-induced performance degradation, consistently outperforming baseline models in both centralized and federated settings. GraMFedDHAR safeguards privacy, boosts performance, and will evolve with larger dataset, adaptive privacy, and smarter fusion.

% \bibliographystyle{unsrt}  


% \end{thebibliography}
% \newpage
\bibliographystyle{unsrt}
\bibliography{Reference}

\begin{thebibliography}{10}

\bibitem{das2020mmhar}
Avigyan Das, Pritam Sil, Pawan~Kumar Singh, Vikrant Bhateja, and Ram Sarkar.
\newblock Mmhar-ensemnet: a multi-modal human activity recognition model.
\newblock {\em IEEE Sensors Journal}, 21(10):11569--11576, 2020.

\bibitem{xiong2022unified}
Baochen Xiong, Xiaoshan Yang, Fan Qi, and Changsheng Xu.
\newblock A unified framework for multi-modal federated learning.
\newblock {\em Neurocomputing}, 480:110--118, 2022.

\bibitem{yang2024cross}
Xiaoshan Yang, Baochen Xiong, Yi~Huang, and Changsheng Xu.
\newblock Cross-modal federated human activity recognition.
\newblock {\em IEEE Transactions on Pattern Analysis and Machine Intelligence}, 46(8):5345--5361, 2024.

\bibitem{garain2022differentially}
Avishek Garain, Rudrajit Dawn, Saswat Singh, and Chandreyee Chowdhury.
\newblock Differentially private human activity recognition for smartphone users.
\newblock {\em Multimedia Tools and Applications}, 81(28):40827--40848, 2022.

\bibitem{abadi2016deep}
Martin Abadi, Andy Chu, Ian Goodfellow, H~Brendan McMahan, Ilya Mironov, Kunal Talwar, and Li~Zhang.
\newblock Deep learning with differential privacy.
\newblock In {\em Proceedings of the 2016 ACM SIGSAC conference on computer and communications security}, pages 308--318, 2016.

\bibitem{el2022differential}
Ahmed El~Ouadrhiri and Ahmed Abdelhadi.
\newblock Differential privacy for deep and federated learning: A survey.
\newblock {\em IEEE access}, 10:22359--22380, 2022.

\bibitem{wei2020federated}
Kang Wei, Jun Li, Ming Ding, Chuan Ma, Howard~H Yang, Farhad Farokhi, Shi Jin, Tony~QS Quek, and H~Vincent Poor.
\newblock Federated learning with differential privacy: Algorithms and performance analysis.
\newblock {\em IEEE transactions on information forensics and security}, 15:3454--3469, 2020.

\bibitem{roy2023temporal}
Debaditya Roy and {\v{S}}ar{\=u}nas Girdzijauskas.
\newblock Temporal differential privacy for human activity recognition.
\newblock In {\em 2023 IEEE 10th International Conference on Data Science and Advanced Analytics (DSAA)}, pages 1--10. IEEE, 2023.

\bibitem{mondal2020new}
Riktim Mondal, Debadyuti Mukherjee, Pawan~Kumar Singh, Vikrant Bhateja, and Ram Sarkar.
\newblock A new framework for smartphone sensor based human activity recognition using graph neural network.
\newblock {\em IEEE Sensors Journal}, 2020.

\bibitem{bandyopadhyay2025mharfedllm}
Asmit Bandyopadhyay, Rohit Basu, Tanmay Sen, and Swagatam Das.
\newblock Mharfedllm: Multimodal human activity recognition using federated large language model.
\newblock {\em arXiv preprint arXiv:2508.01701}, 2025.

\bibitem{wang2023body}
Ziyi Wang, Yihong Chen, Hao Zheng, Meng Liu, and Ping Huang.
\newblock Body rfid skeleton-based human activity recognition using graph convolution neural network.
\newblock {\em IEEE Transactions on Mobile Computing}, 23(6):7301--7317, 2023.

\bibitem{yang2022activity}
Po~Yang, Congmin Yang, Vitaveska Lanfranchi, and Fabio Ciravegna.
\newblock Activity graph based convolutional neural network for human activity recognition using acceleration and gyroscope data.
\newblock {\em IEEE Transactions on Industrial Informatics}, 18(10):6619--6630, 2022.

\bibitem{sarkar2021grafehty}
Abhishek Sarkar, Tanmay Sen, and Ashis~Kumar Roy.
\newblock Grafehty: Graph neural network using federated learning for human activity recognition.
\newblock In {\em 2021 20th IEEE International Conference on Machine Learning and Applications (ICMLA)}, pages 1124--1129. IEEE, 2021.

\bibitem{ahmad2021graph}
Tasweer Ahmad, Lianwen Jin, Xin Zhang, Songxuan Lai, Guozhi Tang, and Luojun Lin.
\newblock Graph convolutional neural network for human action recognition: A comprehensive survey.
\newblock {\em IEEE Transactions on Artificial Intelligence}, 2(2):128--145, 2021.

\bibitem{kulsoom2022review}
Farzana Kulsoom, Sanam Narejo, Zahid Mehmood, Hassan~Nazeer Chaudhry, Ayesha Butt, and Ali~Kashif Bashir.
\newblock A review of machine learning-based human activity recognition for diverse applications.
\newblock {\em Neural Computing and Applications}, 34(21):18289--18324, 2022.

\bibitem{ek2022lightweight}
Sannara Ek, Fran{\c{c}}ois Portet, and Philippe Lalanda.
\newblock Lightweight transformers for human activity recognition on mobile devices.
\newblock {\em arXiv preprint arXiv:2209.11750}, 2022.

\bibitem{zakariyya2025differentially}
Idris Zakariyya, Linda Tran, Kaushik~Bhargav Sivangi, Paul Henderson, and Fani Deligianni.
\newblock Differentially private integrated decision gradients (idg-dp) for radar-based human activity recognition.
\newblock In {\em 2025 IEEE/CVF Winter Conference on Applications of Computer Vision (WACV)}, pages 5611--5622. IEEE, 2025.

\bibitem{subasi2020human}
Abdulhamit Subasi, Kholoud Khateeb, Tayeb Brahimi, and Akila Sarirete.
\newblock Human activity recognition using machine learning methods in a smart healthcare environment.
\newblock In {\em Innovation in health informatics}, pages 123--144. Elsevier, 2020.

\bibitem{singh2020deep}
Satya~P Singh, Madan~Kumar Sharma, Aim{\'e} Lay-Ekuakille, Deepak Gangwar, and Sukrit Gupta.
\newblock Deep convlstm with self-attention for human activity decoding using wearable sensors.
\newblock {\em IEEE Sensors Journal}, 21(6):8575--8582, 2020.

\bibitem{vaswani2017attention}
Ashish Vaswani, Noam Shazeer, Niki Parmar, Jakob Uszkoreit, Llion Jones, Aidan~N Gomez, {\L}ukasz Kaiser, and Illia Polosukhin.
\newblock Attention is all you need.
\newblock In {\em Advances in neural information processing systems}, pages 5998--6008, 2017.

\bibitem{kipf2016semi}
Thomas~N Kipf and Max Welling.
\newblock Semi-supervised classification with graph convolutional networks.
\newblock {\em arXiv preprint arXiv:1609.02907}, 2016.

\bibitem{convergence}
Elsa Rizk and Ali~H. Sayed.
\newblock A graph federated architecture with privacy preserving learning.
\newblock {\em CoRR}, abs/2104.13215, 2021.

\end{thebibliography}

\end{document}